\documentclass[runningheads]{llncs}

\usepackage{accv}

\usepackage{accvabbrv}

\usepackage{graphicx}
\usepackage{booktabs}

\usepackage[accsupp]{axessibility}  

\usepackage[pagebackref,breaklinks,colorlinks,citecolor=accvblue]{hyperref}

\usepackage{orcidlink}

\usepackage{soul,color}
\usepackage{amsmath}
\usepackage{booktabs}
\usepackage{multirow}
\usepackage{algorithm}
\usepackage{algorithmic}
\usepackage{gensymb}

\newcommand{\cF}{\mathcal{F}}

\newcommand{\bM}{\mathbf{M}}

\newcommand{\bp}{\mathbf{p}}

\newcommand{\cW}{\mathcal{W}}

\newcommand{\bx}{\mathbf{x}}

\newcommand{\cX}{\mathcal{X}}
\newcommand{\cN}{\mathcal{N}}

\newcommand{\thickhline}{%
    \noalign {\ifnum 0=`}\fi \hrule height 1pt
    \futurelet \reserved@a \@xhline
}

\begin{document}

\title{Simultaneous Diffusion Sampling for\\ Conditional LiDAR Generation} 


\author{Ryan Faulkner\inst{1} \and
Luke Haub\inst{2} \and
Simon Ratcliffe\inst{2} \and
Anh-Dzung Doan\inst{1} \and
Ian Reid\inst{1} \and
Tat-Jun Chin\inst{1}}

\authorrunning{R.Faulkner et al.}

\institute{Australian Institute for Machine Learning - University of Adelaide \and
Maptek, Glenside, SA 5065, Australia}

\maketitle

\begin{abstract}
  By enabling capturing of 3D point clouds that reflect the geometry of the immediate environment, LiDAR has emerged as a primary sensor for autonomous systems. If a LiDAR scan is too sparse, occluded by obstacles, or too small in range, enhancing the point cloud scan by while respecting the geometry of the scene is useful for downstream tasks. Motivated by the explosive growth of interest in generative methods in vision, conditional LiDAR generation is starting to take off. This paper proposes a novel simultaneous diffusion sampling methodology to generate point clouds conditioned on the 3D structure of the scene as seen from multiple views. The key idea is to impose multi-view geometric constraints on the generation process, exploiting mutual information for enhanced results. Our method begins by recasting the input scan to multiple new viewpoints around the scan, thus creating multiple synthetic LiDAR scans. Then, the synthetic and input LiDAR scans simultaneously undergo conditional generation according to our methodology. Results show that our method can produce accurate and geometrically consistent enhancements to point cloud scans, allowing it to outperform existing methods by a large margin in a variety of benchmarks.
\end{abstract}

\section{Introduction}
\label{sec:intro}

Conditional LiDAR generation (CLG) is a generative process for LiDAR data, which conditions on specific input information to deterministically control desired attributes of the produced output. CLG opens up new possibilities for autonomous systems equipped with LiDAR. For instance, real-world LiDAR scans are often sparse and incomplete, stemming from limited sensor resolution or occlusion; CLG can be used to recover information lost due to partial observations~\cite{VQVAELiDAR}. Another application is LiDAR densification, which produces a high-beam LiDAR scan from a low-beam one~\cite{LiDARGen}. Such capabilities could benefit downstream tasks such as 3D object detection~\cite{pang2020clocs,zhou2020end,fan2021rangedet}, 3D segmentation~\cite{faulkner2023semantic,douillard2011segmentation,zhu2021cylindrical}, or 3D object tracking~\cite{dewan2016motion,chen2023voxelnext,wang2022deepfusionmot}. 

\begin{figure}
    \centering
    \subfloat[][]
    {
        \includegraphics[width=0.3\textwidth]{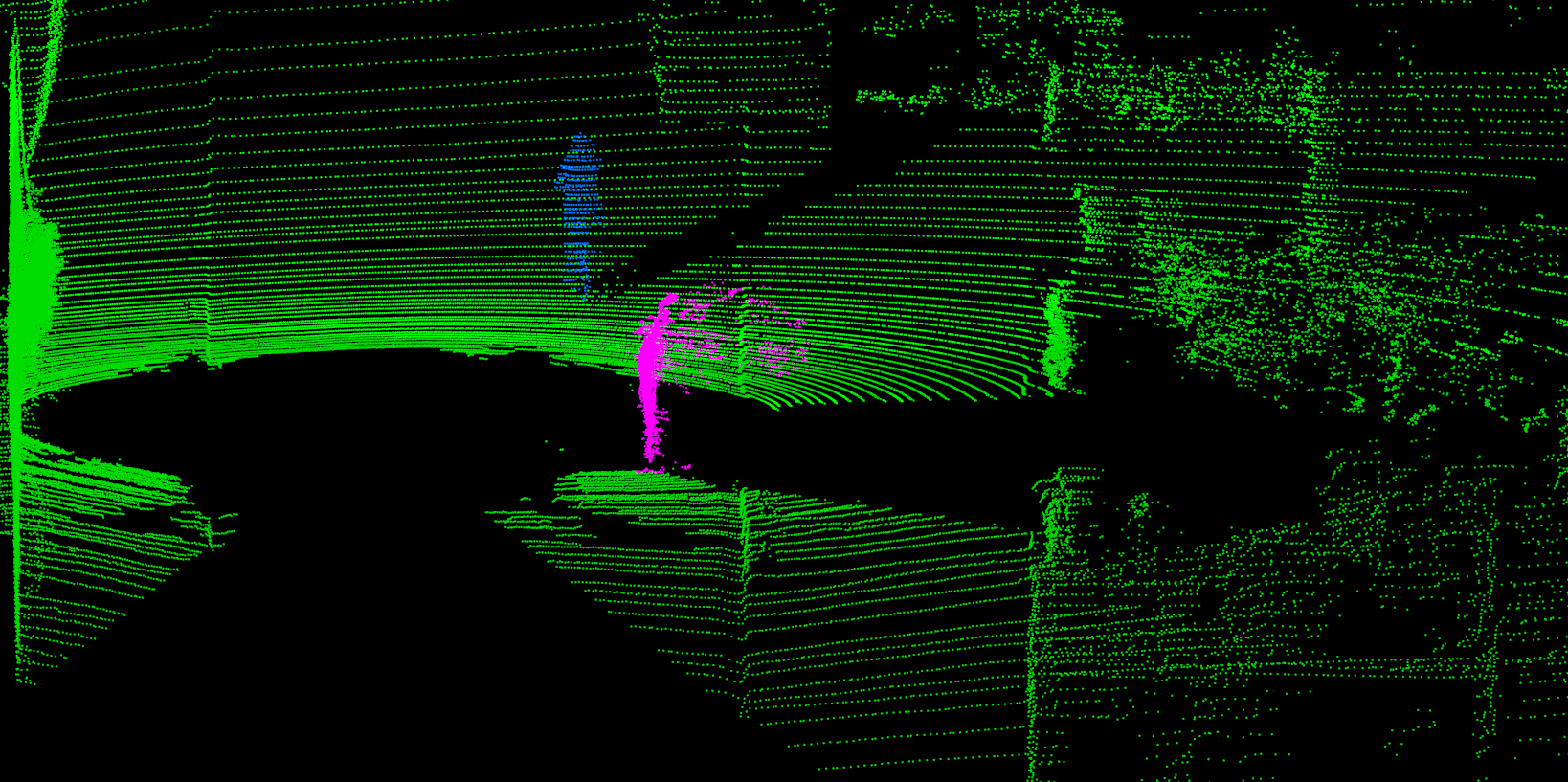}
        
        \label{fig:CompletionInputColour}
    }
    \subfloat[][]
    {
        \includegraphics[width=0.3\textwidth]{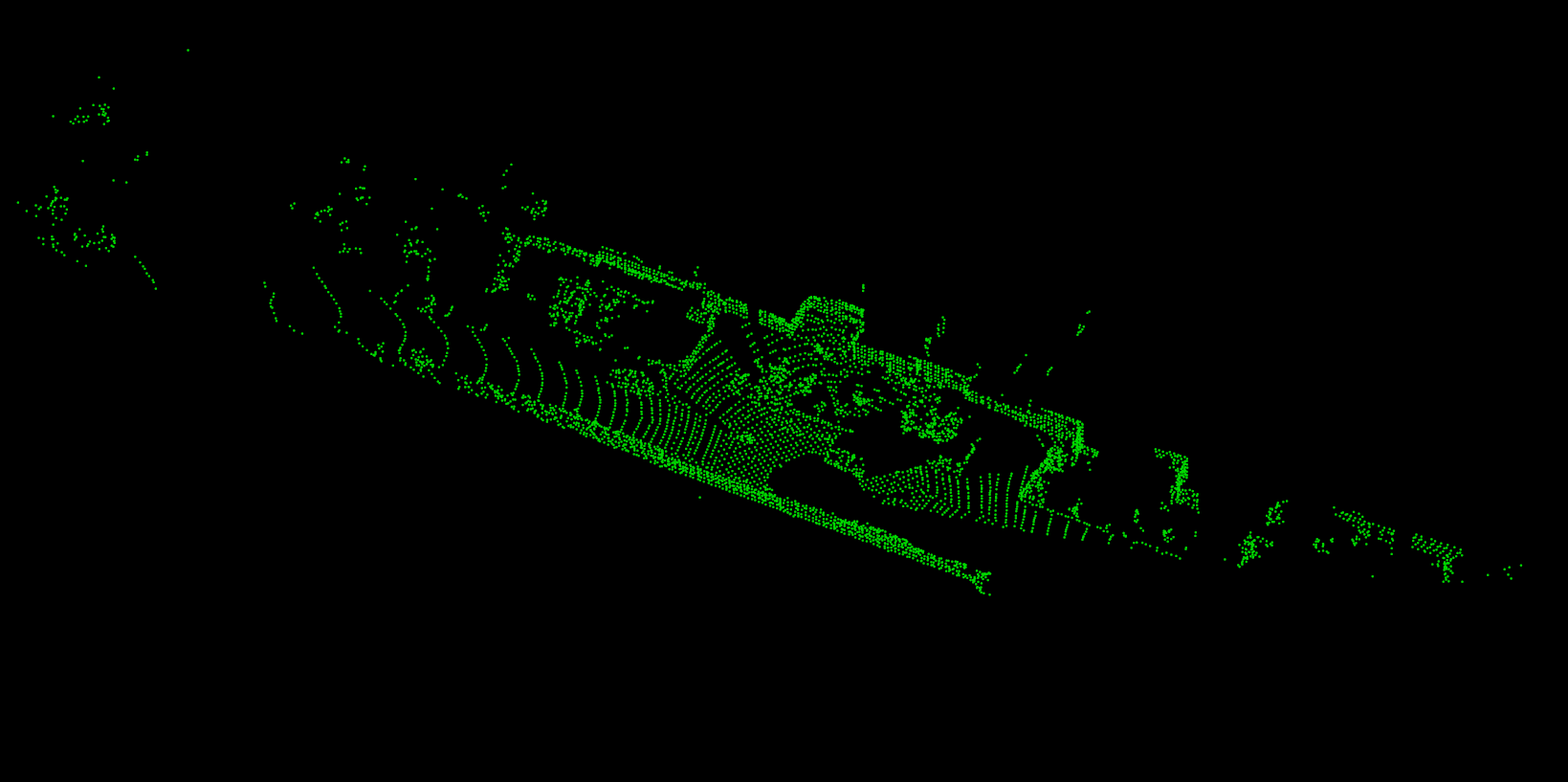}
        
        \label{fig:CompletionInput}
    }
     \subfloat[][]
    {
        \includegraphics[width=0.3\textwidth]{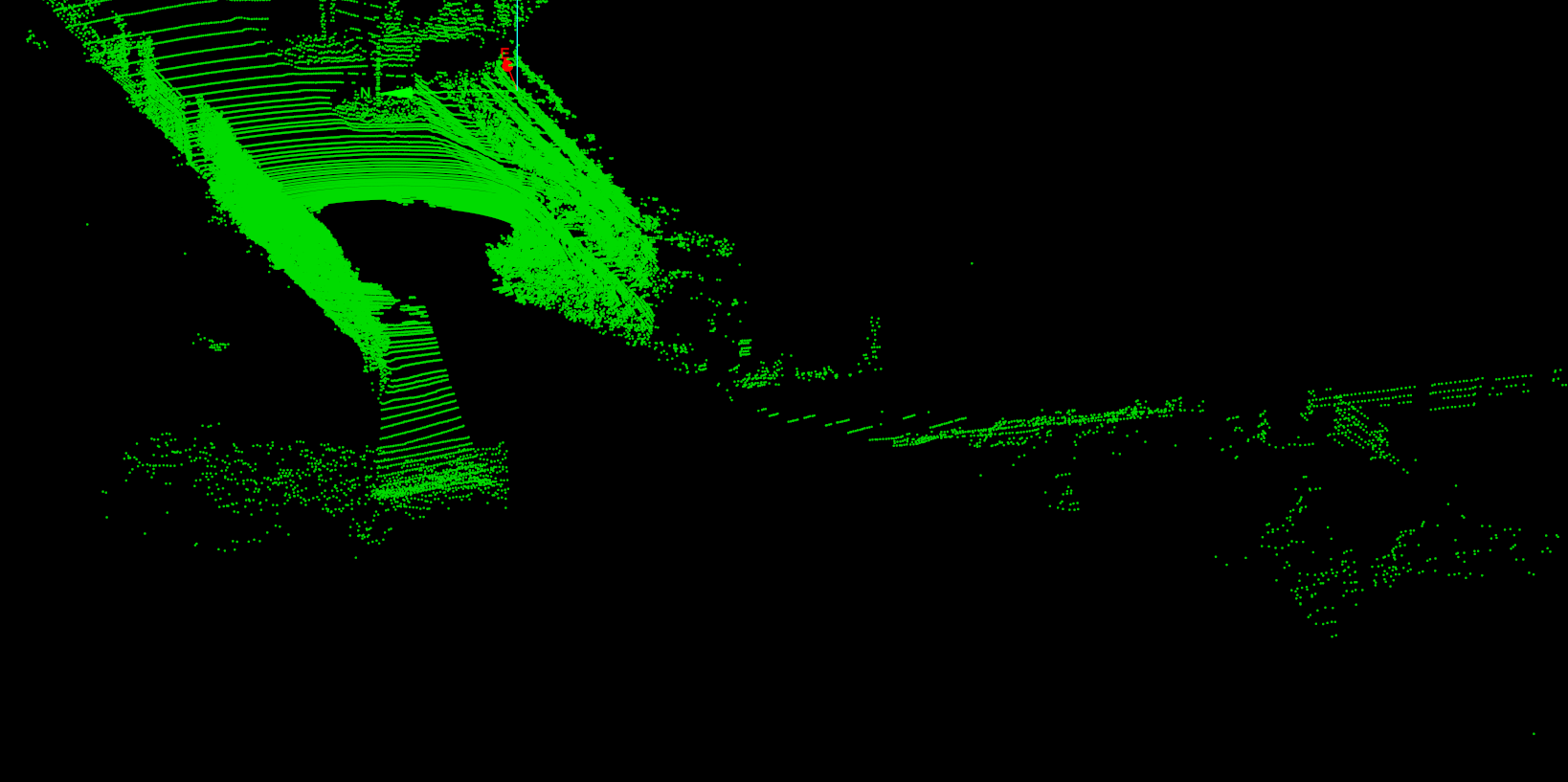}
        
        \label{fig:Before3}
    }\\

    \subfloat[][]
    {
        \includegraphics[width=0.3\textwidth]{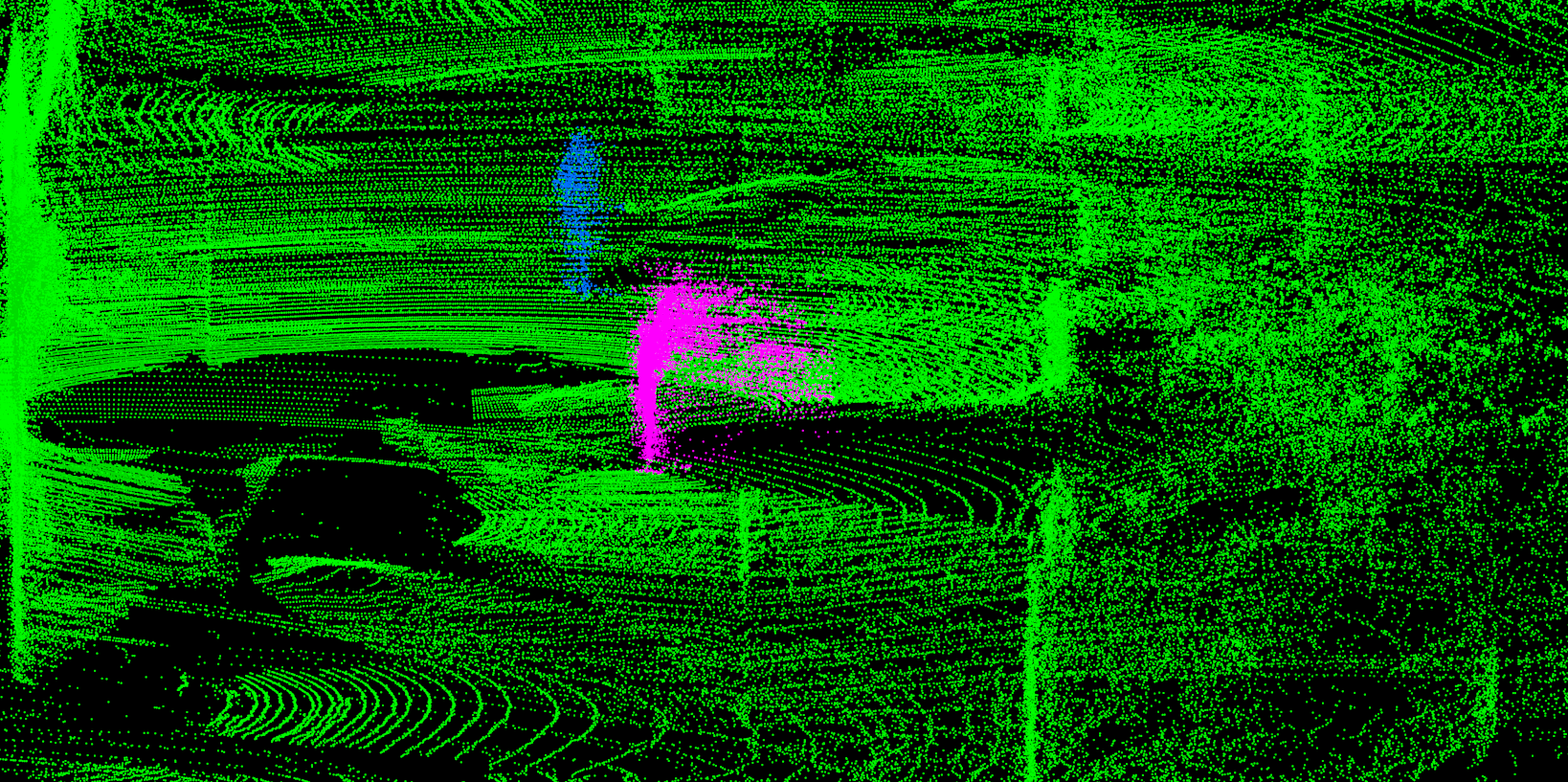}
        
        \label{fig:CompletionOutputColour}
    }
    \subfloat[][]
    {
        \includegraphics[width=0.3\textwidth]{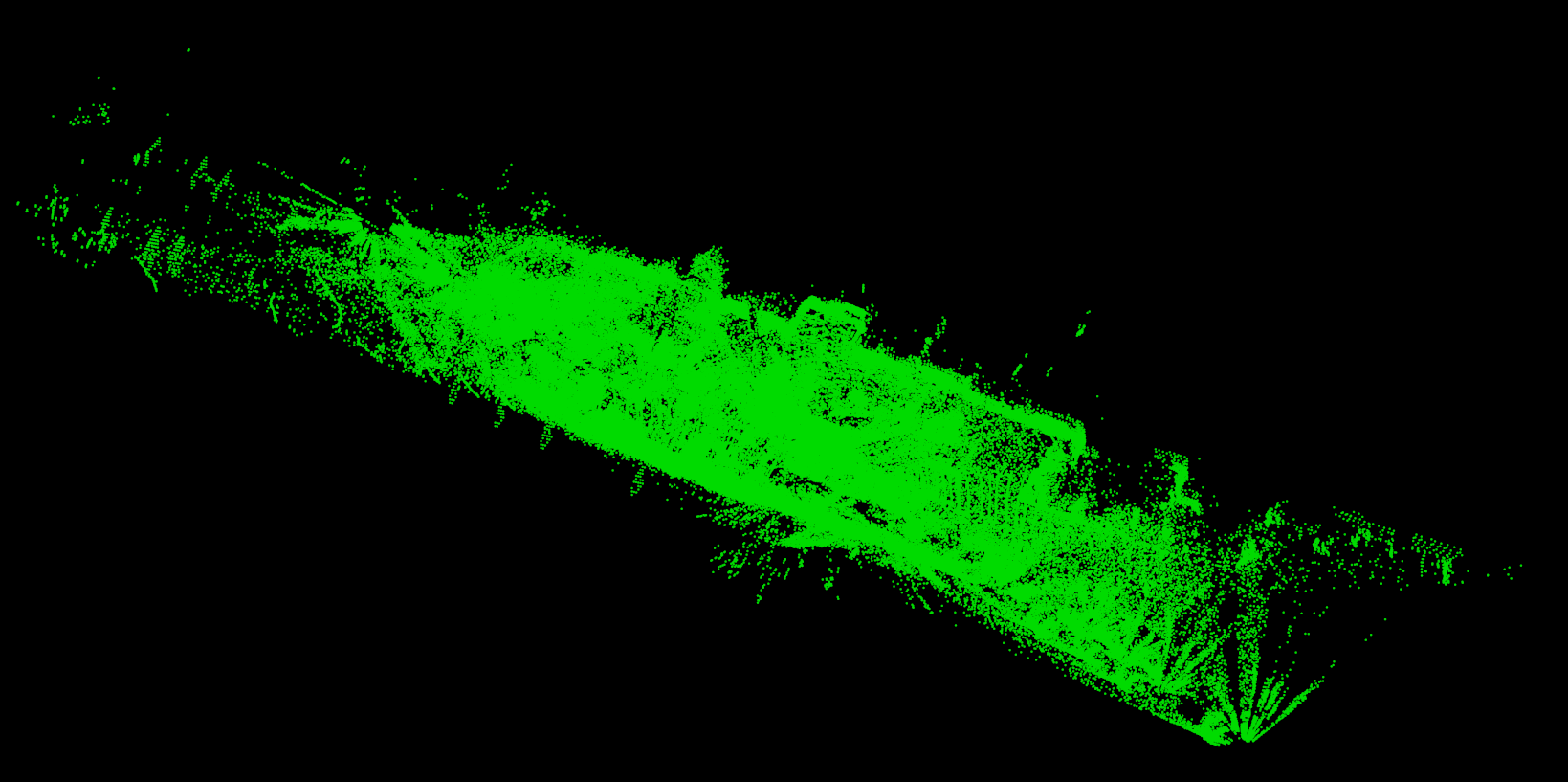}
        
        \label{fig:CompletionOutput}
    }
    \subfloat[][]
    {
        \includegraphics[width=0.3\textwidth]{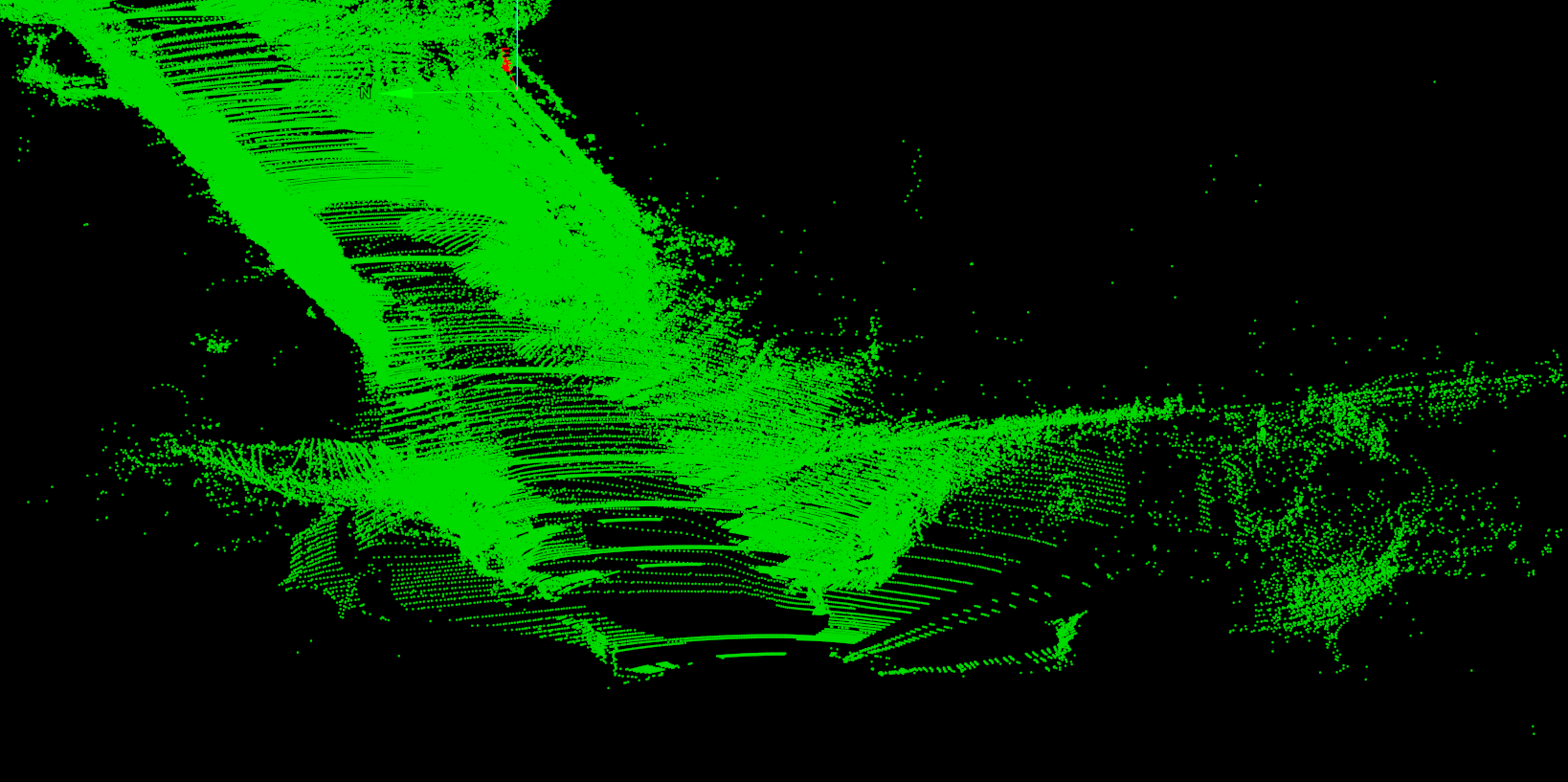}
        
        \label{fig:CompletionAfter4}
    }\\
    \caption{Qualitative results of applying our proposed method to LiDAR scene completion. Top row: the single scan input. Bottom row: Many overlapping synthetic scans, resulting in 50\% more coverage while retaining a high 80\% accuracy. Cyclist and Car coloured in (a,d) for easier readability.}
    \label{fig:sceneCompletionQualitative}    
\vspace{-5pt}
\end{figure}

\begin{figure}
    \centering
    \mbox{
        \subfloat[][]
        {
            \includegraphics[width=0.3\textwidth]{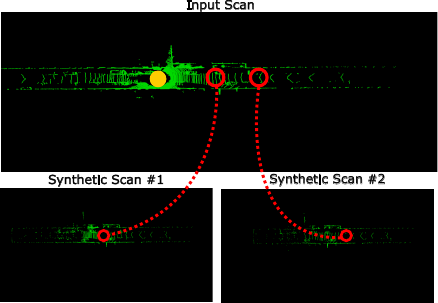}
            
            \label{fig:ProvidedScan}
        }

        \subfloat[][]
        {
            \includegraphics[width=0.3\textwidth]{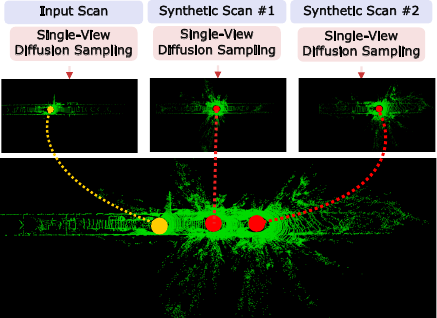}
            
            \label{fig:SingleView}
        }

        \subfloat[][]
        {
            \includegraphics[width=0.3\textwidth]{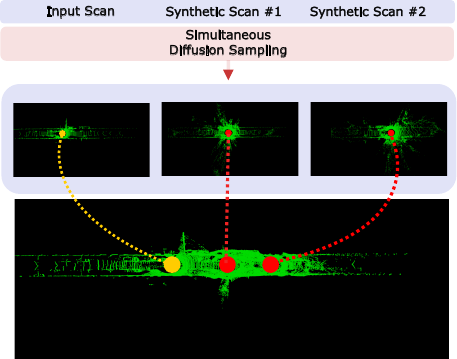}
            
            \label{fig:NovelViewGeneration}
        }
    }
    \caption{Overview of our simultaneous diffusion. The input scan is recasted to generate (partial) synthetic scans. We propose a methodology to apply conditional diffusion sampling to all scans simultaneously, achieving more geometrically consistent results.}
    \label{fig:teasefig}    
    \vspace{-5pt}
\end{figure}

Towards incorporating geometric constraints into CLG to generate higher quality samples, LiDARGen~\cite{LiDARGen} and R2DM~\cite{r2dm} propose to utilise a diffusion model~\cite{diffusionsurvey}. Specifically, a posterior sampling conditioned on the input LiDAR scan is introduced, guaranteeing that the generated LiDAR scan contains a geometric structure similar to that of the input. Although showing promising results in LiDAR densification, other tasks where the pixels to be generated lack nearby reliable groundtruth values (such as inpainting, or novel view generation) can have errors compound as the network struggles to predict noise using nearby pixel values. This is shown for the novel view generation task in \cref{fig:teasefig} where the single view diffusion results in some points ``exploding'' outside of the scene due to compounding errors. A potential approach to better constrain the generation is leveraging geometric consistency across multiple views, which calls for fundamental changes to the diffusion sampling process.

\paragraph{Contributions}

We propose a novel \emph{simultaneous diffusion sampling} technique, that allows conditioning on multiple views of a scene in a way that imposes geometric consistency across all views. To use our technique to enhance an input LiDAR scan, we recast the scan to virtual viewpoints near the scan position, \eg, along a motion trajectory, to create multiple \emph{synthetic} LiDAR scans, which are then subjected to a diffusion sampling constrained by the geometric consistency between the input and synthetic scans; see Fig.~\ref{fig:teasefig}. In this way, we leverage multi-view constraints to enhance the generation of an area of interest, without needing to procure multiple real LiDAR scans. Results show that our technique elevates the CLG performance beyond that achievable by single-view CLG methods, such as LiDARGen~\cite{LiDARGen} and R2DM~\cite{r2dm}, including improving the performance of downstream tasks as well.

As a necessary part of researching and developing this technique, we also explore the application of conditional diffusion sampling to novel view generation for LiDAR scans. We provide analysis of direct geometrical recasting, as well as demonstrate further applications such conditional diffusion sampling of LiDAR scans enables, \eg Scene Completion.

\section{Related work}

\subsection{Generative methods for LiDAR}

There has been significant work in generative models for LiDAR data. While earlier methods utilised GAN and VAE methods~\cite{LiDARVAEandLiDARGAN}, more recent methods have employed diffusion~\cite{LiDARGen,r2dm} and VQ-VAE~\cite{VQVAELiDAR,VQVAEandDiffusionCombo} successfully. 

Specifically, LiDAR-GAN and LiDAR-VAE~\cite{LiDARVAEandLiDARGAN} employ Generative Adversarial Network (GAN)~\cite{gan} and  Variational Autoencoders (VAE)~\cite{kingma2013auto} for CLG. 
However, LiDAR scans generated from LiDAR-GAN and LiDAR-VAE can exhibit noticeable shortcomings in realism. To address this issue, UltraLiDAR~\cite{VQVAELiDAR} leverages vector-quantised VAE~\cite{van2017neural}---a model recognised for its resilience against noise---to produce higher quality point clouds. However, UltraLiDAR conditions its outputs on object categories instead of the scene's geometric structure. To ensure that the generated LiDAR scans are useful for downstream tasks, it is essential to consider the geometry of the scene during the generation process.

Our paper focuses on the more recent application of diffusion models to conditional LiDAR scan generation. While more costly in time compared to other generative methods, diffusion has shown high accuracy and realism in it's synthetic data, and the same trained diffusion model can be easily applied to a wide variety of conditional generation tasks.  



\subsection{Conditional sampling for diffusion models}
While unconditional generation can create synthetic data to expand datasets, many practical applications of diffusion involves conditional generation. Through the use of a given ``condition'' control can be exerted over the otherwise random generation process. Here we distinguish between tasks based on the relationship between the ``condition'' and the desired synthetic data.
\paragraph{Generating with a different data format as the condition}
For these tasks, diffusion is used to effectively convert from one data format to another. Possible formats for both condition and the generated synthetic data include RGB photos, text~\cite{Clip2ImageTrainedWithCondition}, LiDAR scans~\cite{Clip2Image2LiDARPaperOne}, and other representations~\cite{GenerateImagesOfClassesTrainedWithCondition}. Recently Xu \etal ~\cite{dmv3d} used diffusion to generate 3D representations of small objects. They also use multiple views for improved geometry, however their method relies on a transformer-base for which large, complex scenes are beyond it's scope~\cite{hong2024lrm}.

\paragraph{Generating with the same data format as the condition}
When the condition is the same data format and the synthetic data produced, the task is some form of enhancement. Examples include increasing image resolution, inpainting missing sections of an image, or image denoising. This form of conditional generation, specifically enhancing a given LiDAR scan, is the focus of our paper.

While diffusion research often focuses on (unconditional) generative models, we are not the first to focus on conditional generation for data enhancement. There are various conditional sampling methods which achieve high performance by training a different model for each conditional generation task~\cite{ImageResolutionIncreasingTrainedWithCondition,PaletteImageEnhancementTrainedWithCondition}. Recently, Nunes \etal ~\cite{nunes2024cvpr} used diffusion to effectively densify LiDAR scans, determining the nearby area around each existing point. 

Our proposed sampling method is more multi-purpose, sampling using any diffusion model originally trained for unconditional generation as a base, as has been done previously by ~\cite{ScoreMatchingDiffusionPaper,RePaintConditionalDDPM,betterAveragingOfMultipleConditionalSamplingsInARow}. In addition, our sampling method can be applied to a large variety of LiDAR enhancement tasks, including those which extending beyond the existing LiDAR scan's range, growing the scene.

A particularly novel technical aspect of our proposed method compared to existing conditional diffusion methodologies, is how it generates it's own additional conditions. Through sampling multiple views and enforcing geometric consistency across all of them, our method is, to our knowledge, the first work to explore having a (pretrained) diffusion network generate \textit{its own} conditions, simultaneous with respect to the sampling itself, to achieve superior performance.

\subsection{Novel view synthesis}

The position and orientation of a sensor can be referred to as the local frame $\cF$. Predicting what the sensor would return from a specific view $\cF_{target}$ is a challenging task. Accomplishing this however, allows applications such as predicting what will be seen in positions difficult or dangerous to place a sensor in reality ~\cite{novelViewLiDARByCombiningDrive}, or to simply provide a more comprehensive view of a given scene or object.

While there are existing benchmark datasets on novel view synthesis~\cite{KITTI360}, the focus is on generating RGB images, and methods assume a large number of available images for each scene~\cite{Nerflidar,notNerfButUsedManyImagesAsInput} which can be used either for training or inference. Novel view synthesis in the context of (LiDAR scans) is scarce, and existing work similarly assumes a set of scans available for a given scene~\cite{novelViewLiDARByCombiningDrive}. To our knowledge, we are the first to generate from a given LiDAR scan, novel synthetic scans at specified target views.

In contrast to NeRF and similar methods ~\cite{KITTI360,Nerflidar,notNerfButUsedManyImagesAsInput}, diffusion models do not require a large amount of available data for the specific scene or object being generated, but instead can apply learned information from training data within the same domain. For this reason, they have been previously been applied to novel view generation of RGB images, either ignoring the scene geometry~\cite{earlyDiffusionForNovelView} or relying on a network to estimate the transformation necessary between views~\cite{diffusionForSingleView}.

We take advantage of LiDAR scans having accurate geometrical information available to transform data between views both quickly and accurately. A partial synthetic scan for the target view $\cF_{target}$ is synthesised using simple geometry to recast the point cloud's origin. Empty sections of the resultant scan are filled using conditional sampling of a diffusion model. In effect, we approach novel view generation via diffusion as an extension of the existing inpainting task.

\section{Background: Diffusion sampling for CLG}\label{sec:backgroundMethods}


\begin{figure*}
    \centering
    \includegraphics[width=0.9\textwidth]{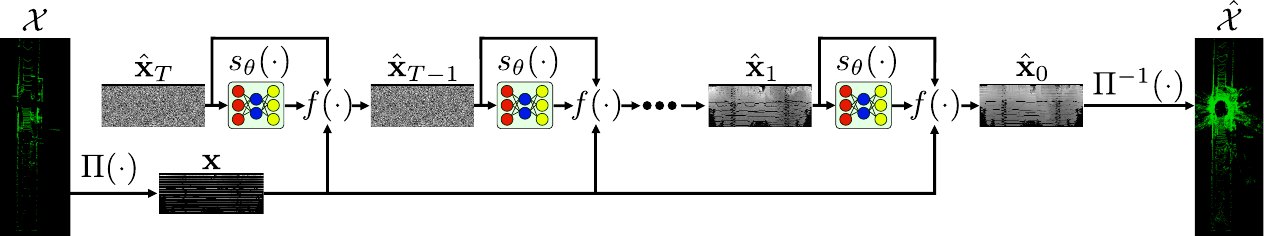}
    \caption{The input LiDAR scan $\cX$ is first projected to the equirectangular image $\bx$. Then, successive sampling conditioned on image $\bx$ is performed to generate the image $\hat{\bx}_0$. Finally, this image is backprojected to obtain the generated LiDAR scan $\hat{\cX}$.}
    \label{fig:lidargen_pipeline}    
\vspace{-5pt}
\end{figure*}

Let an input LiDAR scan of $n$ points be $\cX = \{(\bp_j, r_j)\}_{j=1}^n$, where each point has a 3D position $\bp_j = (x_j, y_j, z_j)$ and a remission $r_j$. The 3D points are expressed in the local frame $\cF$ of the sensor, which relates to a world frame $\cF^W$ via a rigid transformation $\bM^is {W} \in SE(3)$.

First, the scan $\cX$ is spherically projected to an image $\bx$ (~\ref{sec:projections}). Then, a diffusion sampling conditioned on $\bx$ is performed to produce image $\hat{\bx}$ (Sec.~\ref{sec:diffusion}). Finally, the image $\hat{\bx}$ is backprojected to obtain a generated scan $\hat{\cX}$ (Sec.~\ref{sec:projections}). This pipeline is shown in~\cref{fig:lidargen_pipeline} and described in more detail in the following.

\subsection{Spherical projection}\label{sec:projections}

Let $\bx = \Pi(\cX)$ be a spherical projection that projects a LiDAR scan $\cX$ to a two-channel equirectangular image $\bx$ with height $h$ and width $w$. Each 3D point $\bp_j$ is projected to the corresponding 2D pixel coordinate
\begin{align} \label{eq:project:coord}
& \begin{pmatrix}
    u_j \\
    v_j
\end{pmatrix} :=
\begin{pmatrix}
    \left\lfloor\frac{1}{2} \left(1 - \frac{\arctan(\frac{y_j}{x_j})}{\pi} \right)\cdot w \right\rfloor\\
    \left\lfloor\left(1 - \frac{\arcsin\left( \frac{z_j}{d_j} \right)+ f_\text{up}}{f} \right) \cdot h\right\rfloor
\end{pmatrix},
\end{align}
where $f = f_\text{up} + f_\text{down}$ is the vertical field of view of the LiDAR sensor, and $\lfloor \cdot \rfloor$ is the floor function. The pixel $(u_j,v_j)$ contains two elements, the normalised depth and remission, computed as
\begin{align}\label{eq:project:pixel}
\bx(u_j,v_j) := \left(\frac{1}{\alpha}\log_2(d_j+1), \frac{1}{255}r_j \right),
\end{align}
where $d_j = \sqrt{x_j^2 + y_j^2 + z_j^s}$ is the depth, and $\alpha$ is a normalising factor set as necessary to bring the maximum depth to roughly 1. Note that $\alpha$ is empirically set to 6 for automotive datasets such as KITTI-360~\cite{KITTI360}. If multiple 3D points are projected to the same pixel, the point with the smallest depth is used.

Denote by $\cX = \Pi^{-1}(\bx)$ as the inverse spherical projection (``backprojection") that projects $\bx$ back to $\cX$. The derivation of the backprojection is straightforward from Eqs.~\eqref{eq:project:coord} and~\eqref{eq:project:pixel}.

\subsection{Diffusion sampling} \label{sec:diffusion}
Let $s_\theta$ be a score or noise prediction network parameterised by $\theta$. The elaboration of training of $s_\theta$ for CLG can be found in~\cite{LiDARGen} and~\cite{r2dm}. Here, we assume $s_\theta$ has been trained. The diffusion sampling for CLG can be generally defined as

\begin{align}
    \hat{\bx}_{t-1} = f\left(\hat{\bx}_t, s_\theta(\hat{\bx}_t), \bx \right)
\end{align}
where, $f$ can be the Langevin dynamics~\cite{LiDARGen} or reverse diffusion process conditioned on the input $\bx$~\cite{r2dm} and $\hat{\bx}_T \sim \cN(0, I)$.  At the sampling step $t=0$, the image $\hat{\bx}_0$ is projected back to obtain the final output $\hat{\cX} = \Pi^{-1}(\hat{\bx}_0)$.

\section{Simultaneous diffusion sampling for CLG}\label{sec:methodology}
\begin{figure*}
    \centering
    \includegraphics[width=1.0\textwidth]{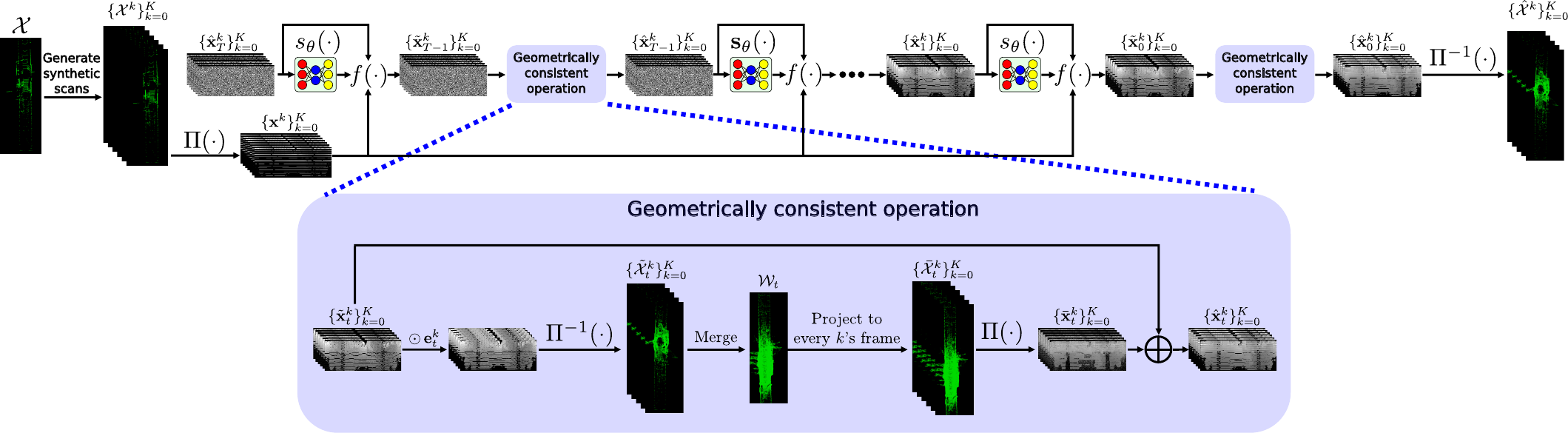}
    \caption{Our proposed method: given the input LiDAR scan $\cX$, we generate a set of $K$ synthetic LiDAR scans from different viewpoints $\{\cX^k\}_{k=0}^K$, where $\cX^0\equiv\cX$. We then project LiDAR scans $\{\cX^k\}_{k=0}^K$ to a set of equirectangular images $\{\bx^k\}_{k=0}^K$ . Next, a diffusion sampling conditioned on images $\{\bx^k\}_{k=0}^K$ is simultaneously performed to generate a set of images $\{\Tilde{\bx}_t^k\}_{k=0}^K$, followed by a geometrically consistent operation to obtain $\{\hat{\bx}_t^k\}_{k=0}^K$, which share a consistent 3D geometry. At the last sampling step $t=0$, we backproject images $\{\hat{\bx}_0^k\}_{k=0}^K$ to LiDAR scans $\{\hat{\cX}^k \}_{k=0}^K$. For tasks with a single output, such as inpainting or densification, only $\hat{\cX}^0$ is returned.}
    \label{fig:our_pipeline}    
\vspace{-5pt}
\end{figure*}

Recall that $s_\theta$ is the noise or noise prediction network. The proposed method begins at the stage where $s_\theta$ has been trained.

Let the input LiDAR scan used for the CLG ``condition'' be now represented as $\cX^0 = \{^0\bp_j, r_j \}_{j=1}^n$, which has the sensor frame $\cF^0$ and corresponding rigid transformation $\bM^W_0$. We propose to generate $K$ synthetic LiDAR scans centred at locations near $\cX^0$, then carry out a diffusion sampling simultaneously on those virtual LiDAR scans conditioned on the geometric consistency between them. Our method is illustrated in \cref{fig:our_pipeline}, and described in detail in the following.

\subsection{Recasting the input LiDAR scan}\label{sec:gen_sensors}

To generate these synthetic scans, their local sensor frames or ``viewpoints'' $\{ \cF^k \}^{K}_{k=1}$, as encoded by the respective local-to-world rigid transformations $\{ \bM^W_k \}^{K}_{k=1}$, are first determined, either being provided as input or using our default settings.

When enhancing a partial input (\eg densification, inpainting, or scene completion) the synthetic scan's origins are either placed around the input scan in a circle, or the road is approximated and scans are placed along it. We explore the effect of different synthetic view placements in \cref{tab:ViewPlacementTable}. 

For the novel view generation task, the synthetic scan's ``pose'' (position and orientation) in world-coordinates are already provided as the targets to be generated. For this purpose, we use the local-to-world transformations provided as part of the KITTI-360 dataset to convert between given scan poses. 


Given the new viewpoints, $K$ synthetic LiDAR scans $\{ \cX^k \}^{K}_{k=1}$ are generated, where $\cX^k = \{^k\bp_j, r_j \}_{j=1}^n$, by recasting $\cX^0$ to each synthetic view. This is achieved by transforming the points from $\cX^0$ to the novel views
\begin{align}
\begin{aligned}
    ^k\tilde{\bp}_j &= (\bM^W_k)^{-1}. \bM^W_0 . ^0\tilde{\bp}_j = \bM^{k}_{0}.^0\tilde{\bp}_j, \;\;\;\; \forall j, k,
\end{aligned}
\end{align}
where $\tilde{\bp} = [\bp^T~1]^T$ is $\bp$ in homogeneous coordinates, and $\bM^{k}_{0}$ is the relative transformation between $\cF^0$ and $\cF^k$. Note the remission $\{ r_j \}^{n}_{j=1}$ for each point is unchanged by the recasting. See Fig.~\ref{fig:ProvidedScan} for examples of LiDAR recasting.

\subsection{Simultaneous diffusion sampling}
\label{sec:method}

 Our method is summarised in Alg.~\ref{algo:simultaneous_sampling}, while details for important steps are below.

Given the $K+1$ LiDAR scans $\{\cX^k\}_{k=0}^K$, we project them to a set $K+1$ of equirectangular images $\{\bx^k\}_{k=0}^K$ using $\Pi$ (line~\ref{algo:line:input_lidar_to_image}). 

At each sampling step $t$, recall that the function $f$ can be the Langevin dynamics or the reverse diffusion process~\cite{LiDARGen, r2dm}; we sample images of the viewpoints $\{\cF^k\}_{k=0}^K$ (line~\ref{algo:line:sample}).
\begin{align}
    \Tilde{\bx}_{t-1} = f\left(\hat{\bx}_t, s_\theta(\hat{\bx}_t), \bx^k \right)
    \label{eq:simul_langevin}
\end{align}
the sampling for each viewpoint being conditioned on its corresponding scan $\bx^k$. 


\begin{algorithm}[h]
    \begin{algorithmic}[1]
        \REQUIRE LiDAR scans $\{\cX^k\}_{k=0}^K$, trained model $s_\theta(\bx)$, contribution ratio $\omega$, and total sampling steps $T$.

        \STATE Project LiDAR scans $\{\cX^k\}_{k=0}^K$ to a set of equirectangular images  $\{\bx^k\}_{k=0}^K$, using $\Pi$ (see Sec.~\ref{sec:projections}). \label{algo:line:input_lidar_to_image}
        
        \STATE Initialise $\hat{\bx}^0_T, \dots, \hat{\bx}^K_T \sim \cN(0, I)$.

        \FOR{$t$ = $T$ to 1}
            \FOR{$k=0$ to $K$}
                \STATE Sample $\Tilde{\bx}^k_{t-1}$ conditioned on $\bx^k$, using Eq.~\eqref{eq:simul_langevin} \label{algo:line:sample}
            \ENDFOR
            \STATE Backproject images $\{\Tilde{\bx}^k_{t-1}\}^K_{k=0}$ to LiDAR scans $\{\Tilde{\cX}^k_{t-1}\}_{k=0}^K$ using $\Pi^{-1}$ (see Sec.~\ref{sec:projections}). \label{algo:line:backproject}
            
            \STATE Discard points in $\{\Tilde{\cX}^k_{t-1}\}_{k=0}^K$ outside scanner limitations

            \STATE Transform 3D points $\{\Tilde{\cX}^k_{t-1}\}_{k=0}^K$ from local sensor frame to world frame, and then merge them together to obtain $\cW_{t-1}$; see Eq.~\eqref{eq:merge}. \label{algo:line:sensor_to_world_then_merge}

            \STATE Transform all 3D points within $\cW_{t-1}$ to every local sensor frames $\{\cF^k\}_{k=0}^K$ to obtain LiDAR scans $\{\bar{\cX}^k_{t-1}\}_{k=0}^K$; see Eq.~\eqref{eq:geo_consistent_lidar}. \label{algo:line:world_to_sensor}
            \STATE Discard points in $\{\bar{\cX}^k_{t-1}\}_{k=0}^K$ outside scanner limitations

            \STATE Project LiDAR scans $\{\bar{\cX}^k_{t-1}\}_{k=0}^K$ to images $\{\bar{\bx}^k_{t-1}\}_{k=0}^K$ using $\Pi$ (see Sec.~\ref{sec:projections}). \label{algo:line:geo_consistent_lidar_to_image}

            \STATE Set pixels in $\{\bar{\bx}^k_{t-1}\}_{k=0}^K$ to equal $\{\Tilde{\bx}^k_{t-1}\}_{k=0}^K$ where difference greater than $\delta$.

            \STATE Linearly combine $\{\Tilde{\bx}^k_{t-1}\}_{k=0}^K$ and $\{\bar{\bx}^k_{t-1}\}_{k=0}^K$ to obtain $\{\hat{\bx}^k_{t-1}\}_{k=0}^K$; see Eq.~\eqref{eq:linear_combine}. \label{algo:line:linear_combine}
        \ENDFOR
        \STATE Backproject images $\{\hat{\bx}^k_0\}_{k=0}^K$ to LiDAR scans $\{\hat{\cX}^k\}_{k=0}^K$ using $\Pi^{-1}$ (see Sec.~\ref{sec:projections}). \label{algo:line:output_image_to_lidar}
        \RETURN Generated LiDAR scans $\{\hat{\cX}^k\}_{k=0}^K$.
    \end{algorithmic}
    \caption{Simultaneous diffusion sampling}
    \label{algo:simultaneous_sampling}
\end{algorithm}

Subsequently, the equirectangular images $\{\Tilde{\bx}^k_{t-1}\}^K_{k=0}$ are projected back to obtain LiDAR scans $\{\Tilde{\cX}^k_{t-1}\}_{k=0}^K$ using $\Pi^{-1}$ (line~\ref{algo:line:backproject}), where $\Tilde{\cX}^k_{t-1}$ contains $\Tilde{n}^k_{t-1}$ number of 3D points relative to local sensor frame $\cF^k$. We note that during this projection from pixel images to the LiDAR scans, points which are outside the scanner's known hardware limitations are discarded. Either being too close, or from pixels the scanner consistently registered no result for in the training data (\eg if the scan does not project to a perfectly rectanglular image).

The number of points are therefore inconsistent between $\{\Tilde{\cX}^k_{t-1}\}_{k=0}^K$, as $\Tilde{n}^k_{t-1}$ varies depending on how many points are discarded. For the implementation, discarded points in $\Tilde{\cX}^k_{t-1}$ are given dummy values and ignored in all future steps.

Next, we transform the 3D points of $\{\Tilde{\cX}^k_{t-1}\}_{k=0}^K$ from local sensor frames to the world frame, then merge them to obtain a set of 3D points $\cW_{t-1}$ (line~\ref{algo:line:sensor_to_world_then_merge}).

\begin{align}
    \cW_{t-1} = \{^W\tilde{\bp}_j \}_{j=1}^{\hat{n}_{t-1}} = \bigcup_{k,j} \left\{\bM^W_k.^k\tilde{\bp}_j\right\}
    \label{eq:merge}     
\end{align}

where $\hat{n}_{t-1} = \sum_k \tilde{n}^k_{t-1}$. Next, all 3D points within the set $\cW_{t-1}$ are transformed to every local sensor frame $\{\cF^k\}_{k=0}^K$ obtaining scans $\{\bar{\cX}^k_{t-1}\}_{k=0}^K$ (line~\ref{algo:line:world_to_sensor}).

\begin{align}
    ^k\tilde{\bp}_j = (\bM^W_k)^{-1}.^W\tilde{\bp}_j, \; \forall k, \; \forall  \, ^W\bp_j \in \cW_t 
    \label{eq:geo_consistent_lidar}
\end{align}
Then, LiDAR scans $\{\bar{\cX}^k_{t-1}\}_{k=0}^K$ are projected to images $\{\bar{\bx}^k_{t-1}\}_{k=0}^K$ using the projection $\Pi$ (line~\ref{algo:line:geo_consistent_lidar_to_image}). Every $\bar{\bx}^k_{t-1}$ is created using 3D points across all viewpoints, ensuring all $\{\bar{\bx}^k_{t-1}\}_{k=0}^K$ are geometrically consistent. 

One danger of creating an image consistent across all views is errors propagating across each scan. To counter this, for pixels where the geometrically consistent image $\{\bar{\bx}^k_{t-1}\}_{k=0}^K$ differs greatly with the original image $\Tilde{\bx}^k_{t-1}$, we default to the original. We use the depth channel to compare the difference between pixels, and set a limit $\delta$ in metres. Pixels in $\{\bar{\bx}^k_{t-1}\}_{k=0}^K$ beyond the limit are set to equal $\Tilde{\bx}^k_{t-1}$. The effect of different values for $\delta$ is explored in the supplementary.

Given two versions of equirectangular images: the initial prediction $\Tilde{\bx}^k_{t-1}$ and the geometrically consistent image $\bar{\bx}^k_{t-1}$, we linearly combine $\{\Tilde{\bx}^k_{t-1}\}_{k=0}^K$ and $\{\bar{\bx}^k_{t-1}\}_{k=0}^K$ to obtain $\{\hat{\bx}^k_{t-1}\}_{k=0}^K$ (line~\ref{algo:line:linear_combine}).
        
\begin{align}
    \hat{\bx}^k_{t-1} = (1-\omega).\Tilde{\bx}^k_{t-1} + \omega.\bar{\bx}^k_{t-1}, \,\, \forall k \in \{0,\dots,K \}
    \label{eq:linear_combine}
\end{align}
where, $\omega$ is a hyperparemeter determining the contribution ratio of $\Tilde{\bx}^k_{t-1}$ and $\bar{\bx}^k_{t-1}$ toward $\hat{\bx}^k_{t-1}$. The effect of different values for $\omega$ is explored in the supplementary.

At the sampling step $t=0$, using $\Pi^{-1}$, the set of $\{\hat{\bx}^k_0\}_{k=0}^K$ are backprojected to obtain the generated LiDAR scans $\{\hat{\cX}^k\}_{k=0}^K$ (line~\ref{algo:line:output_image_to_lidar}).

\section{Results}\label{sec:experiments_results}

\subsection{Experimental setup}


We used the automotive LiDAR dataset KITTI-360~\cite{KITTI360} to evaluate our proposed simultaneous sampling. The dataset includes over 10,000 unique frames (and their corresponding scans), so our computing resources did not allow for testing on the full dataset. Therefore, we randomly sampled testing scans from KITTI-360's ``Drive 0'' subset. Each 64-beam scan has a full 360\degree horizontal field of view, with $f_\text{up} = 3\degree$ and $f_\text{down} = -25\degree$. Scans were projected via $\Pi(\cdot)$ to a 64x1024 image. All methods were tested using the same subset of test scans.

\paragraph{Methods and settings:}


We evaluated across a variety of CLG tasks:

\begin{itemize}
    \item Novel View Generation: Generating a new scan with target frame $\cF_{target}$, given a scan $\cX^0$ with frame $\cF^0$.
    \item Scene Completion: Enhancing a given point cloud in both density and coverage. 
    \item Densification: Generating a 64 beam LiDAR scan given a 16 beam scan $\cX^0$.
    \item Inpainting: For a given gap in the horizontal scope of a scan $\cX^0$, generate values to obtain a 360\degree scan. Due to space constraints these evaluations are in the supplementary.
\end{itemize}

For these tasks, we benchmarked the following pipelines:
\begin{itemize}
    \item LiDARGen~\cite{LiDARGen} with the original ``single-view'' diffusion sampling (Sec.~\ref{sec:backgroundMethods}).
    \item LiDARGen~\cite{LiDARGen} with the proposed simultaneous diffusion sampling (Sec.~\ref{sec:method}).
    \item R2DM~\cite{r2dm} with the proposed simultaneous diffusion sampling (Sec.~\ref{sec:method}).
    \item R2DM~\cite{r2dm} with the original ``single-view'' diffusion sampling (Sec.~\ref{sec:backgroundMethods}).
    \item Navier-Stokes~\cite{NavierStokesInpainting} for the inpainting and novel view generation tasks   (Sec.~\ref{sec:novelviewgenexp} and supplementary).
    \item Bilinear, bicubic and nearest neighbour interpolation for the densification task (Sec.~\ref{sec:densificationexp}).
    \item In addition to comparisons run locally, we submitted our results to a public scene completion benchmark, achieving the best mix of coverage and accuracy. (Sec.~\ref{sec:sceneCompletion}).
\end{itemize}
Note that the last two pipelines are classical (non-learning) methods. 


We employed the pretrained LiDARGen or R2DM diffusion networks for both the single-view diffusion sampling and simultaneous diffusion sampling of each method, following their paper and public code's parameters. These networks were trained using the full KITTI-360 training data, with further details in their respective papers. For our proposed simultaneous diffusion sampling, we set $\omega$ to 0.1 and $\delta$ to 5 m. For Navier-Stokes inpainting, we used a circular neighbourhood with radius of 3. Our simultaneous sampling is only 0.005\% slower for tasks generating multiple scans (novel view, scene completion). Extra scans generated for other tasks (densification, inpainting) increase computation accordingly.


\paragraph{Metrics}

\begin{table*}
    \centering
    \begin{tabular}{lccccccc}
        \toprule
         \multicolumn{8}{c}{Accuracy of Viewpoint Recasting} \\
         \midrule
         Synthetic Origin $k$ & $k=1$ & $k=2$ & $k=3$ & $k=4$ & $k=5$ & $k=6$ & $k=7$\\
         Average Distance from Input View & 4.4 & 8.8 & 13.2 & 17.6 & 22.1 & 26.4 & 30.8 \\
         \% of Input Pixels with Values & 70.6\% & 57.6\% & 50.8\% & 46.5\% & 43.8\% & 39.6\% & 36.4\%\\
         Depth Error & 1.83 & 2.30 & 2.81 & 3.30 & 3.81 & 4.41 & 4.81\\
         Remission Error & 15 & 16.5 & 17.4 & 17.6 & 17.7 & 17.6 & 17.4\\
         \bottomrule
    \end{tabular}
    \caption{Accuracy of our view recasting method detailed in \cref{sec:gen_sensors} compared against groundtruth scans for set viewpoints $k$ positions after the given input scan from the dataset's drive route. As the average distance increases the proportion and accuracy of pixels generated through recasting alone (no diffusion sampling) decreases.}
\label{tab:RecastTable}
\vspace{-5pt}
\end{table*}

\begin{table*}[ht]
    \centering
    \begin{tabular}{lccccccc}
        \toprule
         \multicolumn{8}{c}{Novel View Generation Task $k$} \\
         \midrule
         Synthetic Origin $k$ & $k=1$ & $k=2$ & $k=3$ & $k=4$ & $k=5$ & $k=6$ & $k=7$\\
         \midrule
         \midrule
         \multicolumn{8}{c}{Depth Error} \\
         \midrule
         
         Navier-Stokes & 3.06 & 3.17 & 3.36 & 3.60 & 3.93 & 4.26 & 4.60\\
         
         LiDARGen (Single View) & 2.92 & 3.32 & 3.51 & 3.73 & 3.94 & 4.15 & 4.29\\
          
         LiDARGen (Simultaneous) & 2.60 & 2.85 & 3.11 & 3.24 & 3.39 & 3.47 & \textbf{3.53}\\

         R2DM (Single View) & 2.36 & 2.66 & 2.95 & 3.17 & 3.32 & 3.64 & 3.73 \\

         R2DM (Simultaneous) & \textbf{2.33} & \textbf{2.60} & \textbf{2.84} & \textbf{3.02} & \textbf{3.21} & \textbf{3.44} & 3.78 \\
         \midrule
         \midrule
         \multicolumn{8}{c}{Remission Error } \\
         \midrule
         
         Navier-Stokes & 17.2 & 18.2 & 17.7 & 18.3 & 18.6 & 18.0 & 18.2\\
         
         LiDARGen (Single View) & 14.8 & 16.0 & 16.3 & 16.4 & 16.3 & 15.7 & 15.6\\
         
         LiDARGen (Simultaneous) & 14.3 & 15.0 & 15.3 & 15.4 & 15.5 & 14.9 & 15.1\\

         R2DM (Single View) & 13.7 & 14.6 & 14.8 & 15.0 & 14.9 & 14.9 & \textbf{14.6} \\

         R2DM (Simultaneous)  & \textbf{13.3} & \textbf{14.1} & \textbf{14.4} & \textbf{14.5} & \textbf{14.5} & \textbf{14.4} & 14.7 \\
         \midrule
         \midrule
         \multicolumn{8}{c}{Semantic IoU} \\
         \midrule
         Navier-Stokes & 0.42 & 0.32 & 0.27 & 0.25 & 0.23 & 0.22 & 0.21\\
         
         LiDARGen (Single View) & 0.44 & 0.34 & 0.27 & 0.24 & 0.22 & 0.22 & 0.22\\
         
         LiDARGen (Simultaneous) & 0.45 & 0.40 & 0.34 & 0.30 & 0.28 & 0.26 & 0.24\\
         
         R2DM (Single View) & 0.46 & 0.38 & 0.34 & 0.32 & 0.30 & 0.27 & \textbf{0.28}\\
         
         R2DM (Simultaneous) & \textbf{0.49} & \textbf{0.43} & \textbf{0.37} & \textbf{0.33} & \textbf{0.31} & \textbf{0.28} & 0.27\\
         \bottomrule
    \end{tabular}
    \caption{Results on KITTI360 for novel view generation. Our simultaneous diffusion sampling (detailed in \cref{sec:method}) outperforms existing diffusion and classical methods.}
\label{tab:DistanceAblationTable}
\vspace{-5pt}
\end{table*}

For all tasks we compared the depth and remission values of predicted equirectangular range images $\hat{\bx}$ with the ground truth scan taken at the same location. The mean absolute error (MAE) in depth (metres) and remission (reflectivity of an object in \%) predictions are obtained.

We also measured the success of downstream semantic segmentation on the result of CLG. Segmentations were produced using RangeNet++~\cite{RangeNet++} with the same setup as~\cite{LiDARGen}, including training on SemanticKITTI~\cite{SemanticKITTI}. The segmentation produced by RangeNet++ for the groundtruth KITTI360 scan was used as the target labels. We measured error in semantic prediction with Intersection over Union (IoU), with a higher number reflecting the ability of conditionally generated data to better aid downstream applications as effectively as real data.

\subsubsection{Novel View Generation}\label{sec:novelviewgenexp}

Taking advantage of the depth information inherent in LiDAR scans, our view recasting method detailed in \cref{sec:gen_sensors} enables the application of any method capable of inpainting, to be applied to novel view generation. Through recasting, we generate a partial synthetic scan to then use as  a condition in the sampling processes. We evaluate the accuracy of this recasting in \cref{tab:RecastTable}. Unsurprisingly both accuracy (only comparing pixels with a recasted value) and pixels present in the partial scan decrease as the distance increases. As our novel view generation pipeline assumes that recasted pixels are reliable, the decrease in accuracy can be considered a limiting factor to our approach.

As this is a novel application of CLG, there are no existing benchmarks, with novel view generation traditionally being applied to RGB images. We therefore approached this task in the form of ``predicting what an automotive LiDAR sensor will see further down the road'' using the KITTI-360 dataset. For a given LiDAR scan $\cX^0$ and it's corresponding frame $\cF^0$ we set each target frame $\cF_{target}$ as the frame of a scan taken later in the drive. By doing so, we can compare to a known groundtruth $\cX_{target}$ to measure error.

Following ~\cite{LiDARGen,r2dm}, we used MAE to compare the predictions to the ground truth. See \cref{tab:DistanceAblationTable} for quantitative results, and the supplementary for qualitative.

We set each target $k$ as being five ``frames'' after the prior one, with the input scan being $k=0$. For reference we provide in \cref{tab:RecastTable} further details on the accuracy of each target synthetic view's initial recasted values. 

Our results in \cref{tab:DistanceAblationTable} demonstrate that our proposed simultaneous diffusion sampling produces consistent improvement for the generated $\hat{\bx}^k$, with reduced error in both depth and remission. Furthermore this benefits the downstream application of semantic segmentation, across all synthetic origin distances.

\begin{table}
\centering
\begin{tabular}{lccc}
 \toprule
  & Accuracy  & Completeness  & F1 \\
 \midrule
 Raw Input & \textbf{98.24} & 19.07 & 32.35 \\
 
 EncDec & 41.36 & \textbf{41.23} & 41.29 \\
 
 LiDARGen (Simultaneous) & 71.12 & 29.89 & 42.09 \\
 
 R2DM (Simultaneous) & 80.37 & 29.49 & \textbf{43.15} \\
 \bottomrule
\end{tabular}
\caption{KITTI 360 Scene Completion: Current Public Leaderboard. A practical application of our view recasting (\cref{sec:gen_sensors}) simultaneous sampling. We achieved the highest harmonic mean (F1 score) between accuracy and completeness.}
\label{tab:SceneCompletion}
\vspace{-5pt}
\end{table}

\subsubsection{Scene Completion}\label{sec:sceneCompletion}
To demonstrate the practical applications enabled by our conditional diffusion for novel view generation pipeline, we applied our method to the public scene completion benchmark for KITTI 360. The current public leaderboard is included as \cref{tab:SceneCompletion}. With both LiDARGen and R2DM we were able to achieve a better harmonic mean between accuracy and coverage (F1 score) than existing methods. Specifically, given a single LiDAR scan which makes up 19\% of a known scene, we generated an additional 50\% coverage, while retaining a high accuracy of 80\%. We are the first to apply diffusion to expanding a scene, EncDec being a voxel-based Unet which directly reconstructs the scene~\cite{KITTI360}. 

Our scene coverage (quantified as ``Completeness'') could be further improved with more distant target origins. However with infinite possible locations, accurately identifying where to generate views from (\eg placing synthetic origins along the open street and not inside buildings) is beyond this paper's scope.

\subsubsection{Densification}\label{sec:densificationexp}

 
 
 

\begin{table}[ht]
\centering
\begin{tabular}{lccc}
 \toprule
  & Depth  & Remission  & Semantic \\
  & Error & Error & IoU \\
 \midrule
 Nearest Neighbour & 1.74 & 6.90 & .457 \\
 
 Bicubic & 2.36 & 7.65 & .462 \\
 
 Bilinear & 2.04 & 7.26 & .493 \\
 
 LiDARGen (Single View) &1.78 & 7.60 & \textbf{.536}  \\
 
 R2DM (Single View) & 1.71 & 6.61 & .436   \\
 \midrule
LiDARGen (Simultaneous) & 1.79 & 7.85 & .518\\

R2DM (Simultaneous) & \textbf{1.60} & \textbf{6.46} & .437\\
 \bottomrule
\end{tabular}
\caption{Quantitative results for the common 4x densification benchmark. Our simultaneous sampling improves results across all metrics.}
\label{tab:DensificationTable}
\vspace{-5pt}
\end{table}

\begin{table}
\centering
\begin{tabular}{lccc}
 \toprule
  & Depth  & Remission  & Semantic \\
  & Error & Error & IoU \\
 \midrule
 R2DM Base & 2.01 & 5.15 & 0.59 \\
 
 5m Circle & 2.06 & 4.53 & 0.60 \\
 
 15m Circle & 2.06 & 4.54 & 0.60 \\
 
 5 and 15m Circles & 1.99 & 4.58 & \textbf{0.61}  \\

 Road Estimation & \textbf{1.96} & \textbf{4.51} & \textbf{0.61} \\
 \bottomrule
\end{tabular}
\caption{Results for different placement strategies for synthetic views given the task of inpainting a scan missing one continuous quarter of it's horizontal scope (270$\degree$ not 360$\degree$). These tests were treated as hyper-parameter tuning, so run on separate test data. $\omega$ was set to 0.1, and $\delta$ was set to 1 metre.}
\label{tab:ViewPlacementTable}
\vspace{-5pt}
\end{table}

For each 64 beam LiDAR scan we, like previous works testing for this task \cite{r2dm,LiDARGen}, removed three out of every four beams to produce a 16 beam scan, used conditional generation to produce a new 64 beam scan, then compared the difference using MAE. 

For our simultaneous generation, we ran a simple ablation test to identify how important the placement of synthetic scans in the scene is to improved performance. We compare three trivial variants of placing synthetic views around the input scan in a circle, as well as a simple linear regression based approach to estimating the road in the scene, and placing views along it. From \cref{tab:ViewPlacementTable} we see that estimating the road performs slightly better, so that is how synthetic views were placed for our experiments.

We run additional ablation tests shown in the supplementary, to determine the added benefit of each additional synthetic scan being generated simultaneously, as well as the effect of our method's hyperparemeters $\omega$ and $\delta$.

\begin{figure}
    \centering
    \mbox{
        \subfloat[][]
        {
            \includegraphics[width=0.3\textwidth,height=0.08\textwidth]{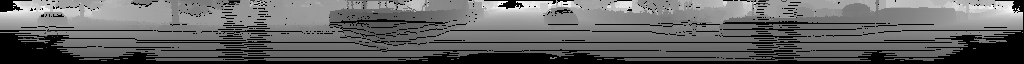}
            \label{fig:GTOutput}
        }

        \subfloat[][]
        {
            \includegraphics[width=0.3\textwidth,height=0.08\textwidth]{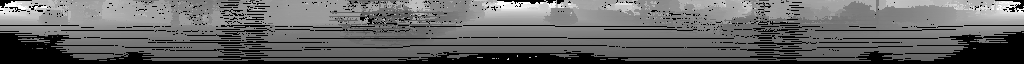}
            \label{fig:LiDARGenOutput}
        }
        \subfloat[][]
        {
            \includegraphics[width=0.3\textwidth,height=0.08\textwidth]{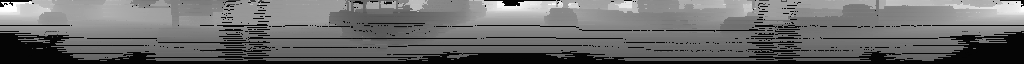}
            \label{fig:R2DMOutput}
        }
    }
    \mbox{
        \subfloat[][]
        {
            \includegraphics[width=0.3\textwidth,height=0.08\textwidth]{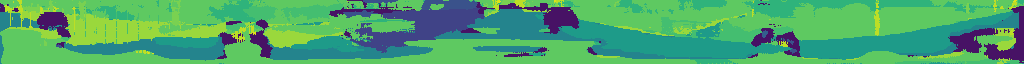}
            \label{fig:GTOutput}
        }

        \subfloat[][]
        {
            \includegraphics[width=0.3\textwidth,height=0.08\textwidth]{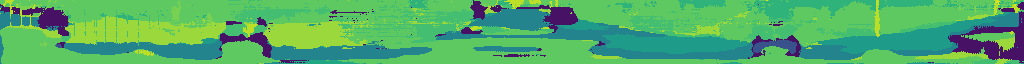}
            \label{fig:LiDARGenOutput}
        }
        \subfloat[][]
        {
            \includegraphics[width=0.3\textwidth,height=0.08\textwidth]{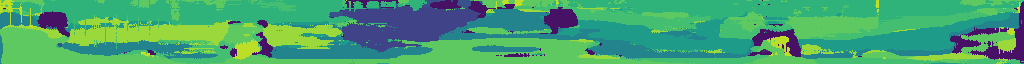}
            \label{fig:R2DMOutputSemantics}
        }
    }
    \caption{Qualitative results for best methods in \cref{tab:DensificationTable}. Top row: depth for Groundtruth, LiDARGen (Single), and R2DM (Simultaneous). Bottom row: semantic prediction.}
    \label{fig:densificationQualitative}    
\end{figure}

As shown in \cref{tab:DensificationTable} our method demonstrates improved depth and remission. The improvement is less significant than other tasks where the input scan does not already have a (sparse) geometry present throughout. Furthermore, our method underperforms on the Semantic IoU metric. As shown in \cref{fig:densificationQualitative} our Simultaneous diffusion generates a noticeably clearer and less noisy result, however that does not result in a similar semantic prediction from the RangeNet++ network, possibly due to noise also being present in the original scans.

\section{Conclusion}
\label{sec:conclusion}

We presented a novel sampling method, applying it to different diffusion models \cite{LiDARGen,r2dm}, we consistently outperform on conditional generation of LiDAR scans compared to the sampling method for models used by existing literature \cite{ScoreMatchingConditionalPaper}.

In addition to this, we demonstrated that conditional sampling for novel view generation can achieve high quality results on public benchmarks in challenging tasks such as scene completion.

\vspace{-0.5em}
\paragraph{Limitations}

We note that a key limitation of our sampling method is the reliance on a depth channel for $\Pi(\cdot)$ and it's inverse function, so cannot be applied to other tasks, such as conditional generation of RGB images. 

Both the recasting and view placement steps also limit our method from being applied to large distances. The recasting process provides the initial synthetic scan for the diffusion, so it's decreased performance with distance becomes a limiting factor. Similarly our approach to scene completion is limited by the ability to place views in reasonable locations. Applying highly accurate diffusion sampling to the generation of a synthetic scan will find limited success if that scan's origin is off the road, inside a building.      

Furthermore, the decreased performance for $k=7$ in \cref{tab:DistanceAblationTable} suggests that when generating synthetic scans in a line (e.g. along a road), the scan furthest away does not benefit from the process.

\vspace{-0.5em}
\paragraph{Further Research}

We demonstrated, to our knowledge for the first time, successful conditional LiDAR generation in the tasks of novel view generation and scene completion. With current literature focused primarily on unconditional generation, we believe there is significant room for further research in areas and methods of applying diffusion architectures.

The experiments performed for this paper either generate synthetic origins further down a known road (\cref{sec:novelviewgenexp}) or when the drive path is unknown (\cref{sec:densificationexp,sec:sceneCompletion}) using a simple road estimator through linear regression. With an infinite number of possible synthetic scan placements, preliminary steps could be developed to identify domain relevant information (such as the road), and from there derive useful scan origins. With infinite possible locations for each novel scan origin, identifying the ``optimal'' placements is a challenging area of research.

Our recasting step could also be an area for improvement. While errors in the depth channel would be difficult to decrease, mainly being from view obstruction affecting what is seen from different positions, the remission channel can clearly be improved. Our current method treats remission as dependant only on the object material, however many factors affect it, such as angle and distance relative to the scanner. Improving the recasting would not only give a better synthetic scan starting point for novel view generation, but also improve the the geometrically consistent image $\bar{\bx}^k_{t-1}$ at every diffusion step of our method.

Finally, in this paper we have implemented and presented the concept of simultaneous diffusion in the context of two diffusion networks, both score and noise predicting. The modification of alternative diffusion pipelines to enable simultaneous sampling is required to confirm that similar improvements and geometric consistency can be achieved for other diffusion architectures.

%
%
\bibliographystyle{splncs04}
\bibliography{main}

\begin{thebibliography}{10}
\providecommand{\url}[1]{\texttt{#1}}
\providecommand{\urlprefix}{URL }
\providecommand{\doi}[1]{https://doi.org/#1}

\bibitem{SemanticKITTI}
Behley, J., Garbade, M., Milioto, A., Quenzel, J., Behnke, S., Stachniss, C., Gall, J.: {SemanticKITTI: A Dataset for Semantic Scene Understanding of LiDAR Sequences}. In: IEEE/CVF International Conference on Computer Vision (2019)

\bibitem{NavierStokesInpainting}
Bertalmio, M., Bertozzi, A., Sapiro, G.: Navier-stokes, fluid dynamics, and image and video inpainting. In: IEEE Conference on Computer Vision and Pattern Recognition (2001)

\bibitem{LiDARVAEandLiDARGAN}
Caccia, L., Van~Hoof, H., Courville, A., Pineau, J.: Deep generative modeling of lidar data. In: IEEE/RSJ International Conference on Intelligent Robots and Systems (2019)

\bibitem{diffusionForSingleView}
Chan, E.R., Nagano, K., Chan, M.A., Bergman, A.W., Park, J.J., Levy, A., Aittala, M., Mello, S.D., Karras, T., Wetzstein, G.: Generative novel view synthesis with 3d-aware diffusion models. arXiv preprint arXiv:2304.02602  (2023)

\bibitem{chen2023voxelnext}
Chen, Y., Liu, J., Zhang, X., Qi, X., Jia, J.: Voxelnext: Fully sparse voxelnet for 3d object detection and tracking. In: IEEE/CVF Conference on Computer Vision and Pattern Recognition (2023)

\bibitem{diffusionsurvey}
Croitoru, F.A., Hondru, V., Ionescu, R.T., Shah, M.: Diffusion models in vision: A survey. IEEE Transactions on Pattern Analysis and Machine Intelligence  (2023)

\bibitem{dewan2016motion}
Dewan, A., Caselitz, T., Tipaldi, G.D., Burgard, W.: Motion-based detection and tracking in 3d lidar scans. In: IEEE International Conference on Robotics and Automation (2016)

\bibitem{GenerateImagesOfClassesTrainedWithCondition}
Dhariwal, P., Nichol, A.: Diffusion models beat gans on image synthesis. In: Advances in Neural Information Processing Systems (2021)

\bibitem{douillard2011segmentation}
Douillard, B., Underwood, J., Kuntz, N., Vlaskine, V., Quadros, A., Morton, P., Frenkel, A.: On the segmentation of {3D} {LIDAR} point clouds. In: IEEE International Conference on Robotics and Automation (2011)

\bibitem{fan2021rangedet}
Fan, L., Xiong, X., Wang, F., Wang, N., Zhang, Z.: {RangeDet}: In defense of range view for lidar-based {3D} object detection. In: IEEE/CVF International Conference on Computer Vision (2021)

\bibitem{faulkner2023semantic}
Faulkner, R., Haub, L., Ratcliffe, S., Reid, I., Chin, T.J.: Semantic segmentation on {3D} point clouds with high density variations. In: 2023 International Conference on Digital Image Computing: Techniques and Applications (2023)

\bibitem{gan}
Goodfellow, I., Pouget-Abadie, J., Mirza, M., Xu, B., Warde-Farley, D., Ozair, S., Courville, A., Bengio, Y.: Generative adversarial nets. In: Advances in Neural Information Processing Systems (2014)

\bibitem{hong2024lrm}
Hong, Y., Zhang, K., Gu, J., Bi, S., Zhou, Y., Liu, D., Liu, F., Sunkavalli, K., Bui, T., Tan, H.: {LRM}: Large reconstruction model for single image to 3d. In: The Twelfth International Conference on Learning Representations (2024), \url{https://openreview.net/forum?id=sllU8vvsFF}

\bibitem{kingma2013auto}
Kingma, D.P., Welling, M.: Auto-encoding variational bayes. arXiv preprint arXiv:1312.6114  (2013)

\bibitem{VQVAEandDiffusionCombo}
Lee, J., Im, W., Lee, S., Yoon, S.E.: Diffusion probabilistic models for scene-scale 3d categorical data. arXiv preprint arXiv:2301.00527  (2023)

\bibitem{KITTI360}
Liao, Y., Xie, J., Geiger, A.: {KITTI}-360: A novel dataset and benchmarks for urban scene understanding in {2D} and {3D}. IEEE Transactions on Pattern Analysis and Machine Intelligence  (2022)

\bibitem{RePaintConditionalDDPM}
Lugmayr, A., Danelljan, M., Romero, A., Yu, F., Timofte, R., Gool, L.V.: Repaint: Inpainting using denoising diffusion probabilistic models. In: IEEE/CVF Conference on Computer Vision and Pattern Recognition (2022)

\bibitem{RangeNet++}
Milioto, A., Vizzo, I., Behley, J., Stachniss, C.: {RangeNet++: Fast and Accurate LiDAR Semantic Segmentation}. In: IEEE/RSJ International Conference on Intelligent Robots and Systems (2019)

\bibitem{r2dm}
Nakashima, K., Kurazume, R.: Lidar data synthesis with denoising diffusion probabilistic models. In: IEEE International Conference on Robotics and Automation (2024)

\bibitem{nunes2024cvpr}
Nunes, L., Marcuzzi, R., Mersch, B., Behley, J., Stachniss, C.: {Scaling Diffusion Models to Real-World 3D LiDAR Scene Completion}. In: {Proc. of the IEEE/CVF Conf. on Computer Vision and Pattern Recognition (CVPR)} (2024)

\bibitem{pang2020clocs}
Pang, S., Morris, D., Radha, H.: {CLOCs}: Camera-{LiDAR} object candidates fusion for {3D} object detection. In: IEEE/RSJ International Conference on Intelligent Robots and Systems (2020)

\bibitem{Clip2ImageTrainedWithCondition}
Ramesh, A., Dhariwal, P., Nichol, A., Chu, C., Chen, M.: Hierarchical text-conditional image generation with clip latents. arXiv preprint arXiv:2204.06125  (2022)

\bibitem{notNerfButUsedManyImagesAsInput}
Riegler, G., Koltun, V.: Free view synthesis. In: European Conference on Computer Vision (2020)

\bibitem{PaletteImageEnhancementTrainedWithCondition}
Saharia, C., Chan, W., Chang, H., Lee, C.A., Ho, J., Salimans, T., Fleet, D.J., Norouzi, M.: Palette: Image-to-image diffusion models. In: ACM SIGGRAPH (2022)

\bibitem{ImageResolutionIncreasingTrainedWithCondition}
Saharia, C., Ho, J., Chan, W., Salimans, T., Fleet, D.J., Norouzi, M.: Image super-resolution via iterative refinement. IEEE Transactions on Pattern Analysis and Machine Intelligence  (2022)

\bibitem{novelViewLiDARByCombiningDrive}
Schmidt, J., Khan, Q., Cremers, D.: Lidar view synthesis for robust vehicle navigation without expert labels. In: IEEE International Conference on Intelligent Transportation Systems (2023)

\bibitem{ScoreMatchingDiffusionPaper}
Song, Y., Ermon, S.: Improved techniques for training score-based generative models. In: Advances in Neural Information Processing Systems (2020)

\bibitem{ScoreMatchingConditionalPaper}
Song, Y., Sohl{-}Dickstein, J., Kingma, D.P., Kumar, A., Ermon, S., Poole, B.: Score-based generative modeling through stochastic differential equations. In: International Conference on Learning Representations (2021)

\bibitem{van2017neural}
Van Den~Oord, A., Vinyals, O., et~al.: Neural discrete representation learning. In: Advances in Neural Information Processing Systems (2017)

\bibitem{wang2022deepfusionmot}
Wang, X., Fu, C., Li, Z., Lai, Y., He, J.: Deepfusionmot: A 3d multi-object tracking framework based on camera-lidar fusion with deep association. IEEE Robotics and Automation Letters  (2022)

\bibitem{earlyDiffusionForNovelView}
Watson, D., Chan, W., Martin-Brualla, R., Ho, J., Tagliasacchi, A., Norou, M.: Novel view synthesis with diffusion models. arXiv preprint arXiv:2210.04628  (2022)

\bibitem{betterAveragingOfMultipleConditionalSamplingsInARow}
Wu, L., Trippe, B.L., Naesseth, C.A., Blei, D.M., Cunningham, J.P.: Practical and asymptotically exact conditional sampling in diffusion models. In: Advances in Neural Information Processing Systems (2024)

\bibitem{VQVAELiDAR}
Xiong, Y., Ma, W., Wang, J., Urtasun, R.: Learning compact representations for lidar completion and generation. In: IEEE/CVF Conference on Computer Vision and Pattern Recognition (2023)

\bibitem{dmv3d}
Xu, Y., Tan, H., Luan, F., Bi, S., Wang, P., Li, J., Shi, Z., Sunkavalli, K., Wetzstein, G., Xu, Z., Zhang, K.: {DMV}3d: Denoising multi-view diffusion using 3d large reconstruction model. In: The Twelfth International Conference on Learning Representations (2024), \url{https://openreview.net/forum?id=H4yQefeXhp}

\bibitem{Clip2Image2LiDARPaperOne}
Xue, L., Gao, M., Xing, C., Mart{\'\i}n-Mart{\'\i}n, R., Wu, J., Xiong, C., Xu, R., Niebles, J.C., Savarese, S.: {ULIP}: Learning unified representation of language, image and point cloud for {3D} understanding. In: IEEE/CVF Conference on Computer Vision and Pattern Recognition (2023)

\bibitem{Nerflidar}
Zhang, J., Zhang, F., Kuang, S., Zhang, L.: Nerf-lidar: Generating realistic lidar point clouds with neural radiance fields. In: AAAI Conference on Artificial Intelligence (2024)

\bibitem{zhou2020end}
Zhou, Y., Sun, P., Zhang, Y., Anguelov, D., Gao, J., Ouyang, T., Guo, J., Ngiam, J., Vasudevan, V.: End-to-end multi-view fusion for {3D} object detection in lidar point clouds. In: Conference on Robot Learning (2020)

\bibitem{zhu2021cylindrical}
Zhu, X., Zhou, H., Wang, T., Hong, F., Ma, Y., Li, W., Li, H., Lin, D.: Cylindrical and asymmetrical 3d convolution networks for lidar segmentation. In: IEEE/CVF conference on computer vision and pattern recognition (2021)

\bibitem{LiDARGen}
Zyrianov, V., Zhu, X., Wang, S.: Learning to generate realistic lidar point clouds. In: European Conference on Computer Vision (2022)

\end{thebibliography}
\title{Supplementary Material for ``Simultaneous Diffusion Sampling for Conditional LiDAR Generation''}

\maketitle
\section{Ablation Tests}
\subsection{Parameter Ablation Tests}\label{sec:inpaintingexp}
We ran some simple ablation tests on the effects of our method's two parameters - $\omega$ and $\delta$, which determined the values used for the tests in our main paper. As shown by \cref{tab:OmegaAblationTable}, higher $\omega$ values improves depth channel performance, however remission is negatively affected. Improving the recasting method's handling of the remission channel would likely lead to higher omega values being optimal. In \cref{tab:deltaTable} we see that having some limiter in place helps prevent errors propagating across the simultaneous sampling. If one view incorrectly places an object \eg extremely close to the scanner, $\delta$ prevents that error being passed on to other views. With that in mind, $\delta$ cannot be too small, or the simultaneous method will provide a negligible benefit / effect.

\begin{table*}
    \centering
    \begin{tabular}{lccccccc}
        \toprule
         \multicolumn{8}{c}{Results for different values of $\omega$} in novel view generation\\
         \midrule
         Synthetic Origin $k$ & $k=1$ & $k=2$ & $k=3$ & $k=4$ & $k=5$ & $k=6$ & $k=7$\\
         \midrule
         \multicolumn{8}{c}{Depth Error \big\downarrow} \\
         \midrule
         $\omega$ = 0 & 2.46 & 2.82 & 3.09 & 3.24 & 3.38 & \textbf{3.48} & \textbf{3.68} \\
         $\omega$ = 0.001 & 2.46 & 2.81 & 3.05 & 3.20 & 3.43 & 3.52 & 3.69 \\
         $\omega$ = 0.01 & 2.46 & 2.81 & 3.05 & 3.20 & 3.43 & 3.51 & 3.69 \\
         $\omega$ = 0.1 & \textbf{2.45} & 2.79 & 3.04 & 3.19 & 3.41 & 3.51 & 3.70 \\
         $\omega$ = 0.2 & \textbf{2.45} & 2.78 & 3.04 & 3.17 & 3.41 & 3.51 & 3.71 \\
         $\omega$ = 0.35 & \textbf{2.45} & 2.78 & 3.00 & 3.16 & 3.38 & 3.49 & 3.73 \\
         $\omega$ = 0.5 & \textbf{2.45} & 2.78 & 3.00 & 3.15 & 3.38 & 3.50 & 3.76 \\
         $\omega$ = 0.65 & \textbf{2.45} & \textbf{2.77} & \textbf{2.98} & \textbf{3.14} & 3.37 & 3.50 & 3.78 \\
         $\omega$ = 0.8 & 2.46 & 2.78 & 3.00 & 3.14 & \textbf{3.36} & 3.52 & 3.80 \\
         $\omega$ = 1 & 2.47 & 2.78 & 3.00 & 3.16 & 3.37 & \textbf{3.49} & 3.84 \\
         \midrule
         \multicolumn{8}{c}{Remission Error \big\downarrow} \\
         $\omega$ = 0 & 13.5 & 14.5 & 14.8 & 14.8 & 14.6 & 14.3 & 14.4 \\
         $\omega$ = 0.001 & 13.3 & 14.2 & 14.6 & 14.4 & 14.6 & 14.4 & 14.2 \\
         $\omega$ = 0.01 & 13.3 & 14.2 & 14.6 & 14.4 & 14.5 & 14.3 & 14.2 \\
         $\omega$ = 0.1 &  \textbf{13.2} & \textbf{14.1} & \textbf{14.5} & \textbf{14.2} & \textbf{14.4} & \textbf{14.1} & \textbf{14.1} \\
         $\omega$ = 0.2 & \textbf{13.2} & \textbf{14.1} & \textbf{14.5} & 14.3 & \textbf{14.4} & 14.2 & 14.1 \\
         $\omega$ = 0.35 & 13.3 & 14.2 & 14.6 & 14.4 & 14.5 & 14.3 & 14.3 \\
         $\omega$ = 0.5 & 13.4 & 14.3 & 14.7 & 14.5 & 14.6 & 14.4 & 14.4 \\
         $\omega$ = 0.65 & 13.4 & 14.3 & 14.7 & 14.6 & 14.7 & 14.5 & 14.5 \\
         $\omega$ = 0.8 & 13.5 & 14.4 & 14.8 & 14.7 & 14.7 & 14.5 & 14.6 \\
         $\omega$ = 1 & 13.6 & 14.6 & 15.1 & 15.0 & 15.1 & 14.8 & 14.9 \\
         \midrule
         \multicolumn{8}{c}{Semantic IoU  \big\uparrow} \\
         $\omega$ = 0 & .362 & .282 & .263 & .247 & .238 & .229 & .214 \\
         $\omega$ = 0.001 & .385 & .312 & .278 & \textbf{.260} & .249 & .234 & .218\\
         $\omega$ = 0.01 & .386 & .316 & .279 & .258 & .250 & \textbf{.236} & \textbf{.219}\\
         $\omega$ = 0.1 & \textbf{.387} & .317 & .283 & \textbf{.260} & \textbf{.252} & .233 & \textbf{.219} \\
         $\omega$ = 0.2 & \textbf{.387} & \textbf{.319} & \textbf{.287} & .256 & .249 & .233 & .216 \\
         $\omega$ = 0.35 & .382 & .318 & \textbf{.287} & .258 & .246 & .229 & .207 \\
         $\omega$ = 0.5 & .379 & .310 & .281 & .253 & .241 & .219 & .207 \\
         $\omega$ = 0.65 & .373 & .300 & .270 & .250 & .236 & .213 & .201 \\
         $\omega$ = 0.8 & .361 & .285 & .258 & .237 & .228 & .206 & .194 \\
         $\omega$ = 1 & .347 & .264 & .239 & .217 & .211 & .195 & .184 \\
         \bottomrule
    \end{tabular}
    \caption{Ablation test of different $\omega$ values. Run generating 7 views for known future locations simultaneously, $\delta$ set to 1 metre.}
\label{tab:OmegaAblationTable}
\end{table*}

\begin{table*}
    \centering
    \begin{tabular}{lccccccc}
        \toprule
         \multicolumn{8}{c}{Results for different values of $\omega$} in novel view generation\\
         \midrule
         Synthetic Origin $k$ & $k=1$ & $k=2$ & $k=3$ & $k=4$ & $k=5$ & $k=6$ & $k=7$\\
         \midrule
         \multicolumn{8}{c}{Depth Error \big\downarrow} \\
         \midrule
         $\delta$ = 0.5 & 2.28 & 2.53 & 2.78 & 3.02 & 3.24 & 3.27 & 3.58 \\
         $\delta$ = 1 & 2.28 & 2.52 & 2.77 & 3.00 & 3.24 & 3.27 & 3.54 \\
         $\delta$ = 2 & 2.28 & 2.50 & 2.76 & 3.00 & 3.21 & 3.22 & 3.56 \\
         $\delta$ = 5 & 2.26 & 2.47 & \textbf{2.72} & 2.92 & 3.11 & 3.21 & 3.45 \\
         $\delta$ = 10 & 2.25 & 2.44 & \textbf{2.72} & \textbf{2.91} & \textbf{3.08} & \textbf{3.17} & \textbf{3.35} \\
        $\delta$ = 15 & 2.26 & 2.45 & \textbf{2.72} & 2.95 & 3.12 & 3.21 & 3.37 \\
         $\delta$ = 20 & 2.26 & \textbf{2.43} & 2.73 & 2.96 & 3.17 & 3.22 & 3.37 \\
         $\delta$ = 30 & 2.24 & 2.46 & 2.75 & 2.96 & 3.20 & 3.28 & 3.40 \\
         $\delta$ = No Limit & \textbf{2.22} & 2.49 & 2.84 & 3.02 & 3.32 & 3.35 & 3.49 \\
         R2DM Base & 2.26 & 2.55 & 2.84 & 3.11 & 3.19 & 3.27 & 3.45 \\
         \midrule
         \multicolumn{8}{c}{Remission Error \big\downarrow} \\
         $\delta$ = 0.5 & 13.1 & 13.8 & 14.5 & \textbf{14.3} & \textbf{14.5} & \textbf{14.2} & 14.2 \\
         $\delta$ = 1 & 13.1 & 13.7 & 14.5 & \textbf{14.3} & \textbf{14.5} & \textbf{14.2} & 14.2 \\
         $\delta$ = 2 & 13.1 & 13.7 & 14.5 & \textbf{14.3} & \textbf{14.5} & 14.2 & 14.3 \\
         $\delta$ = 5 & \textbf{13.0} & \textbf{13.6} & \textbf{14.3} & \textbf{14.3} & \textbf{14.5} & \textbf{14.1} & 14.3 \\
         $\delta$ = 10 & \textbf{13.0} & \textbf{13.6} & 14.4 & 14.4 & 14.6 & 14.2 & 14.6 \\
         $\delta$ = 15 & \textbf{13.0} & 13.7 & 14.4 & \textbf{14.3} & 14.6 & 14.2 & 14.5 \\
         $\delta$ = 20 & \textbf{13.0} & 13.7 & 14.5 & 14.4 & 14.6 & 14.2 & 14.5 \\
         $\delta$ = 30 & \textbf{13.0} & 13.8 & 14.5 & 14.4 & 14.6 & 14.2 & 14.5 \\
         $\delta$ = No Limit & \textbf{13.0} & 13.9 & 14.5 & 14.3 & 14.5 & \textbf{14.1} & 14.4 \\
         R2DM Base & 13.3 & 14.2 & 14.8 & 14.7 & 14.6 & 14.4 & 14.4 \\
         \midrule
         \multicolumn{8}{c}{Semantic IoU  \big\uparrow} \\
         $\delta$ = 0.5 & .390 & .303 & .259 & .233 & \textbf{.236} & .229 & \textbf{.222} \\
         $\delta$ = 1 & .391 & .300 & .258 & .231 & .231 & .228 & .219 \\
         $\delta$ = 2 & \textbf{.392} & .306 & \textbf{.268} & \textbf{.232} & .233 & .230 & .219 \\
         $\delta$ = 5 & \textbf{.392} & \textbf{.308} & .266 & \textbf{.237} & .234 & .232 & .216 \\
         $\delta$ = 10 & .384 & .301 & .262 & .231 & .230 & .219 & .203 \\
         $\delta$ = 15 & .384 & .295 & .261 & .233 & .221 & .210 & .202  \\
         $\delta$ = 20 & .384 & .293 & .259 & .235 & .221 & .213 & .202 \\
         $\delta$ = 30 & .382 & .292 & .260 & .235 & .223 & .213 & .193 \\
         $\delta$ = No Limit & .377 & .296 & .244 & .213 & .193 & .196 & .176 \\
         R2DM Base & .362 & .281 & .245 & .230 & .232 & .234 & .216 \\
         \bottomrule
    \end{tabular}
    \caption{Ablation test of different $\delta$ values. Run generating 7 synthetic views at known future scanner positions simultaneously, accuracy compared against groundtruth scans for those positions. $\omega$ set to 0.1}
\label{tab:deltaTable}
\end{table*}

\subsection{Further Ablation Tests}\label{sec:furtherAblationTests}

\begin{table}
\centering
\begin{tabular}{lccc}
 \toprule
  & Depth & Remission & Semantic \\
  & Error & Error & IoU \\
 \midrule
 \multicolumn{3}{c}{Ten Percent Missing} \\
 \midrule
 Navier-Stokes & 1.39 & 2.44 & .631  \\
 
 R2DM \\ (Single View)& \textbf{1.03} & \textbf{1.80} & \textbf{.725} \\
 \midrule
 R2DM \\ (Simultaneous) & 1.14 & 1.82 & .719 \\
 \midrule
 \multicolumn{3}{c}{Twenty Percent Missing} \\
 \midrule
 Navier-Stokes & 1.65 & 4.17 & .561  \\
 
 R2DM \\ (Single View)& \textbf{1.31} & 2.99 & .647 \\
 \midrule
 R2DM \\ (Simultaneous) & 1.41 & \textbf{2.81} & \textbf{.651} \\
 \midrule
 \multicolumn{3}{c}{Thirty Percent Missing} \\
 \midrule
 Navier-Stokes & 1.99 & 6.20 & .491  \\
 
 R2DM \\ (Single View)& \textbf{1.77} & 4.28 & .567 \\
 \midrule
 R2DM \\ (Simultaneous) & 1.86 & \textbf{4.04} & \textbf{.575} \\
 \midrule
 \multicolumn{3}{c}{Fourty Percent Missing} \\
 \midrule
 Navier-Stokes & 2.41 & 7.83 & .432  \\
 
 R2DM \\ (Single View)& \textbf{2.08} & 5.45 & .511 \\
 \midrule
 R2DM \\ (Simultaneous) & 2.27 & \textbf{4.96} & \textbf{.515} \\
 \midrule
 \multicolumn{3}{c}{Fifty Percent Missing} \\
 \midrule
 Navier-Stokes & 3.46 & 9.31 & .372  \\
 
 R2DM \\ (Single View)& \textbf{2.37} & 6.39 & .451  \\
 \midrule
 R2DM \\ (Simultaneous) & 2.65 & \textbf{5.82}  & \textbf{.460} \\
 \bottomrule
 
\end{tabular}
\caption{Quantitative results for the inpainting task. Our simultaneous method demonstrates improved performance for remission error and semantic IoU across all inpainting variants}
\label{tab:InpaintingTable}
\end{table}

\begin{table}
\centering
\begin{tabular}{lccc}
 \toprule
  & Depth & Remission & Semantic \\
  & Error \big\downarrow & Error \big\downarrow & IoU \big\uparrow \\
 \midrule
 
 Single View & 1.77 & 4.28 & .567 \\
 \midrule
 Simultaneous: 2 Views & 1.81 & 4.14 & .570 \\
 \midrule
 Simultaneous: 3 Views & 2.00 & \textbf{4.05} & .573 \\
 \midrule
 Simultaneous: 4 Views & \textbf{1.75} & 4.07 & .573 \\
 \midrule
 Simultaneous: 5 Views & 1.82 & 4.10 & .571 \\
 \midrule
 Simultaneous: 6 Views & 1.88 & \textbf{4.05} & \textbf{.575} \\
 \midrule
 Simultaneous: 7 Views & 1.86 & 4.09 & \textbf{.575} \\
 \bottomrule
\end{tabular}
\caption{Ablation test for number of origins for the inpainting task. The road in each scene was estimated through linear regression, and views were progressively added 5m ahead, 5m behind, 10m ahead, 10m behind, 15m ahead, 15m behind.}
\label{tab:InpaintingTableOriginAblation}
\end{table}

\begin{table}
\centering
\begin{tabular}{lcccc}
 \toprule
  & View Added & Depth & Remission & Semantic \\
  & & Error \big\downarrow & Error \big\downarrow & IoU \big\uparrow \\
 \midrule
 
 Single View & -- & 2.52 & 11.2 & .317 \\
 \midrule
 Simultaneous: 2 Views & $k=3$  & 2.99 & 11.2 & .288 \\
 \midrule
 Simultaneous: 3 Views & $k=5$ & 2.96 & 11.1 & .285 \\
 \midrule
 Simultaneous: 4 Views & $k=2$ & 2.81 & 11.0 & .287 \\
 \midrule
 Simultaneous: 5 Views & $k=6$ & 2.80 & 10.9 & .289 \\
 \midrule
 Simultaneous: 6 Views & $k=1$ & 2.38 & 10.7 & \textbf{.334} \\
 \midrule
 Simultaneous: 7 Views & $k=7$ & \textbf{2.37} & \textbf{10.6} & .331 \\
 \bottomrule
\end{tabular}
\caption{Ablation test for number of origins for the novel view generation task. Synthetic views were added using known road positions from other scans in the dataset.}
\label{tab:ViewGenerationTableOriginAblation}
\end{table}

We ran a series of ablation tests to further examine the details of our proposed method and benefits it provides. We compared inpainting results across different sized field-of-view gaps. We defined each test by the proportion of image missing, the missing area's centre consistently set to 72\degree to the right of the vehicle's direction. with results shown in \cref{tab:InpaintingTable}. Our sampling method outperforms the default consistently for remission and semantic IoU, however results in worse depth error across all inpainting tests. 

This demonstrates the danger of errors propagating from the more distant (and thus error-prone) viewpoints during the simultaneous sampling process, something also shown when we ran an ablation on the effect of each additional viewpoint, for the thirty percent missing inpainting task. \cref{tab:InpaintingTableOriginAblation} shows the depth error is best with four synthetic views being sampled simultaneously to support the inpainting, the addition of subsequent, more distant views hindering the depth prediction. 

We performed yet another test, comparing the results for a different number origins in the novel view generation task as well. For this set of tests we compared results generating the synthetic view $k=4$, adding the supporting synthetic views in the order of 3,5,2,6,1,7. This set of ablation tests, provided in \cref{tab:ViewGenerationTableOriginAblation}, demonstrates clearly that adding synthetic views between the original input scan $k=0$ and the target $k=4$ provides significant benefit, helping to better utilise the known information from $k=0$. In contrast adding scans further away ($k=5$, $k=6$,$k=7$) provides marginal benefit, simply helping enforce geometric consistency from an extra viewpoint. The requirement of $k=1$ for simultaneous diffusion to outperform the standard single-view diffusion suggests that synthetic views need a minimum amount of information after the recasting (or put another way, a minimum accuracy of the final synthetic scan) to be useful to the overall process. 

Determining the optimal placement of synthetic views, as well as the accuracy required for a synthetic view to be beneficial to generating the overall scene, are both potential areas for further research.

\section{Qualitative Data}\label{sec:Qualitative}
Here we provide some more qualitative results, visualising how our simultaneous method results in more geometrically consistent results. Cars are less likely to merge into nearby walls, generated walls connect to existing ones, and so forth.
\begin{figure}
    \centering
    \subfloat[][]
    {
        \includegraphics[width=0.45\textwidth]{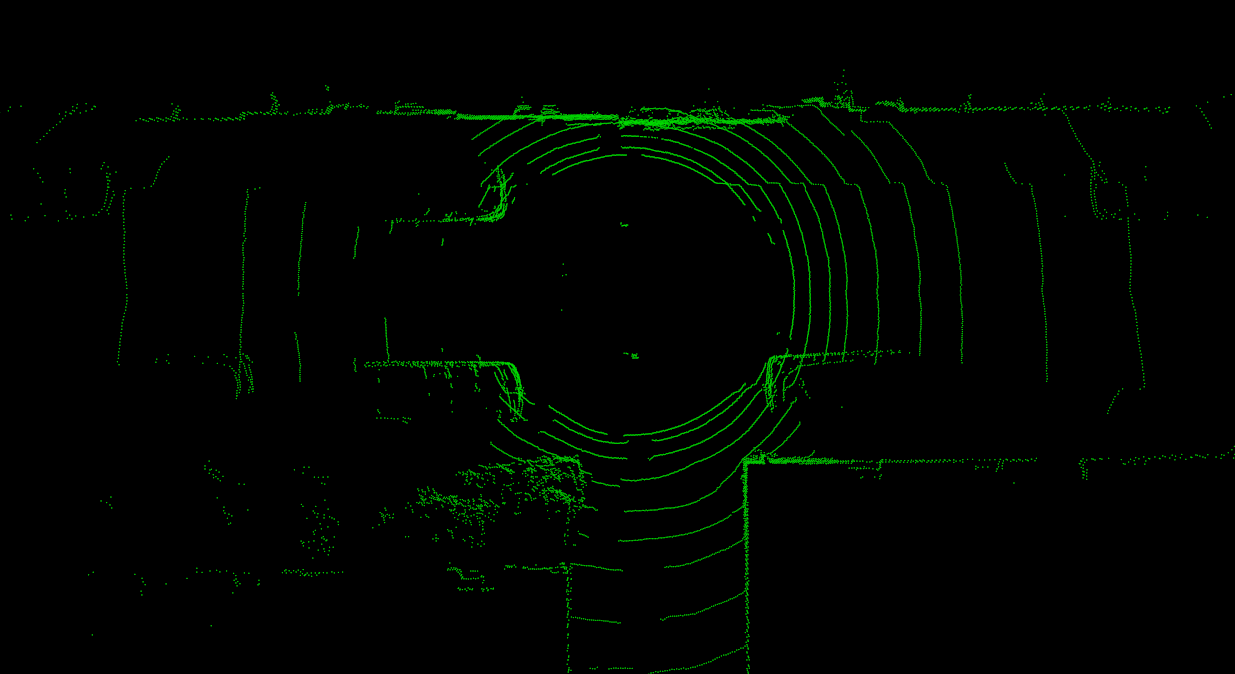}
        
    }
    \subfloat[][]
    {
        \includegraphics[width=0.45\textwidth]{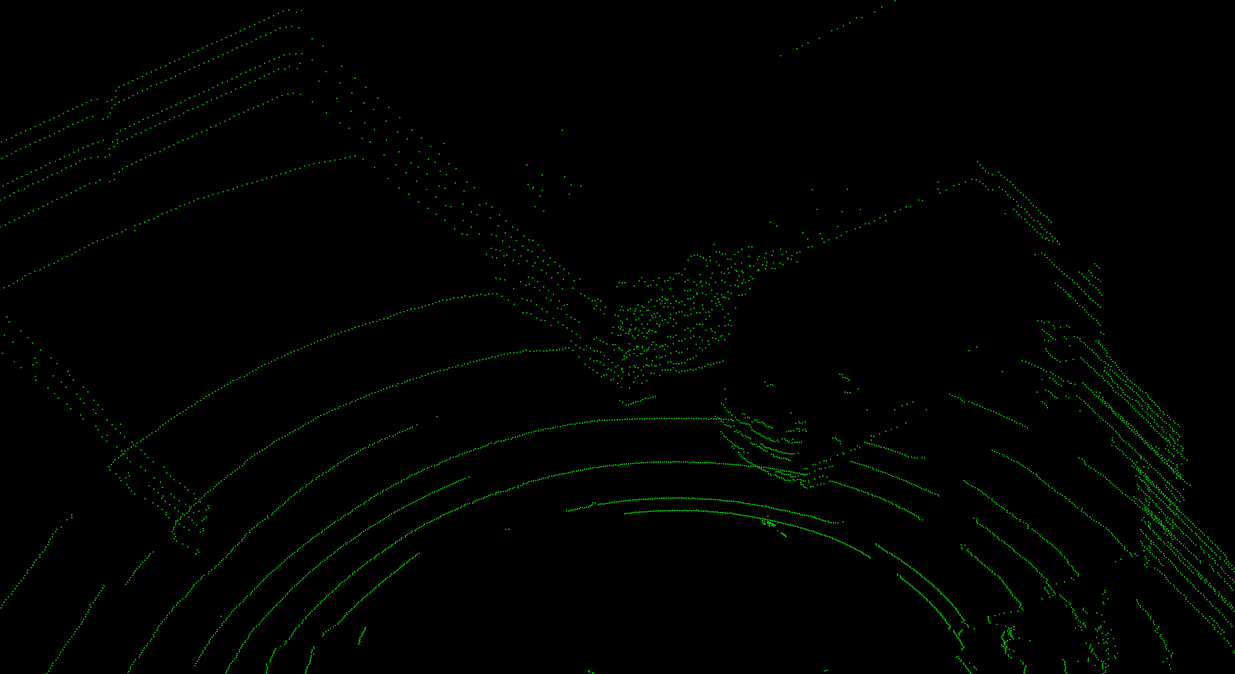}
        
    }\\
    \subfloat[][]
    {
        \includegraphics[width=0.45\textwidth]{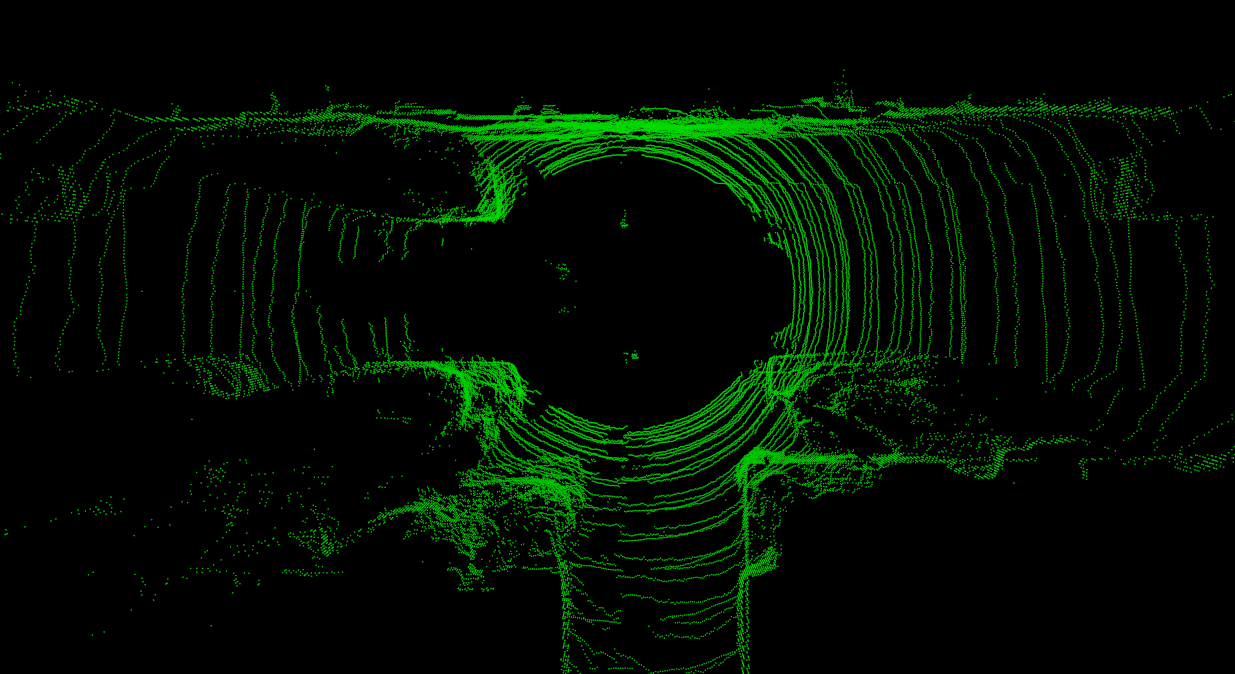}
        
    }
    \subfloat[][]
    {
        \includegraphics[width=0.45\textwidth]{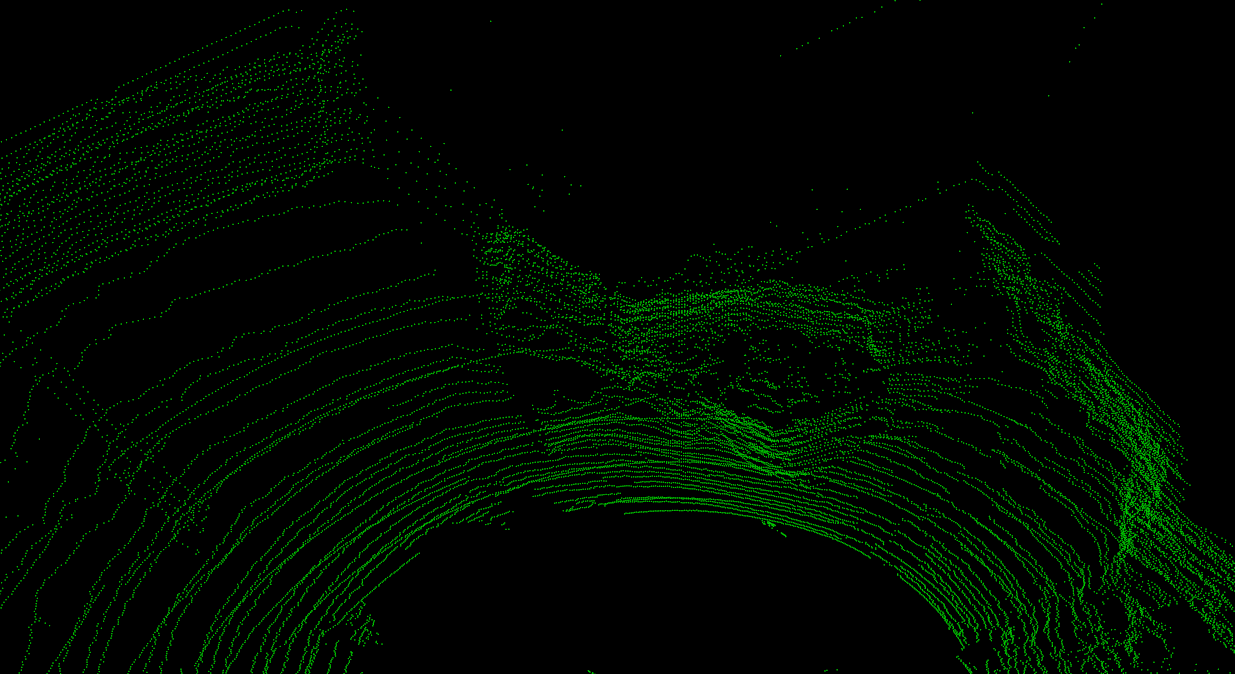}
        
    }\\
     \subfloat[][]
    {
        \includegraphics[width=0.45\textwidth]{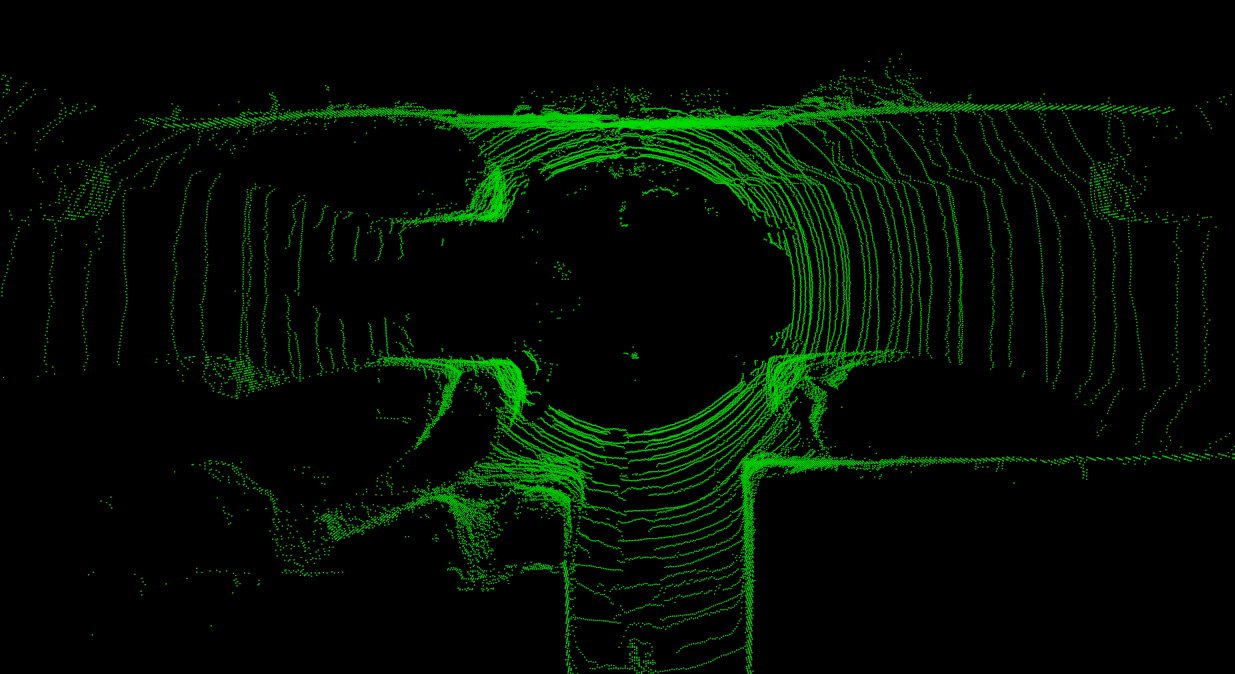}
        
    }
    \subfloat[][]
    {
        \includegraphics[width=0.45\textwidth]{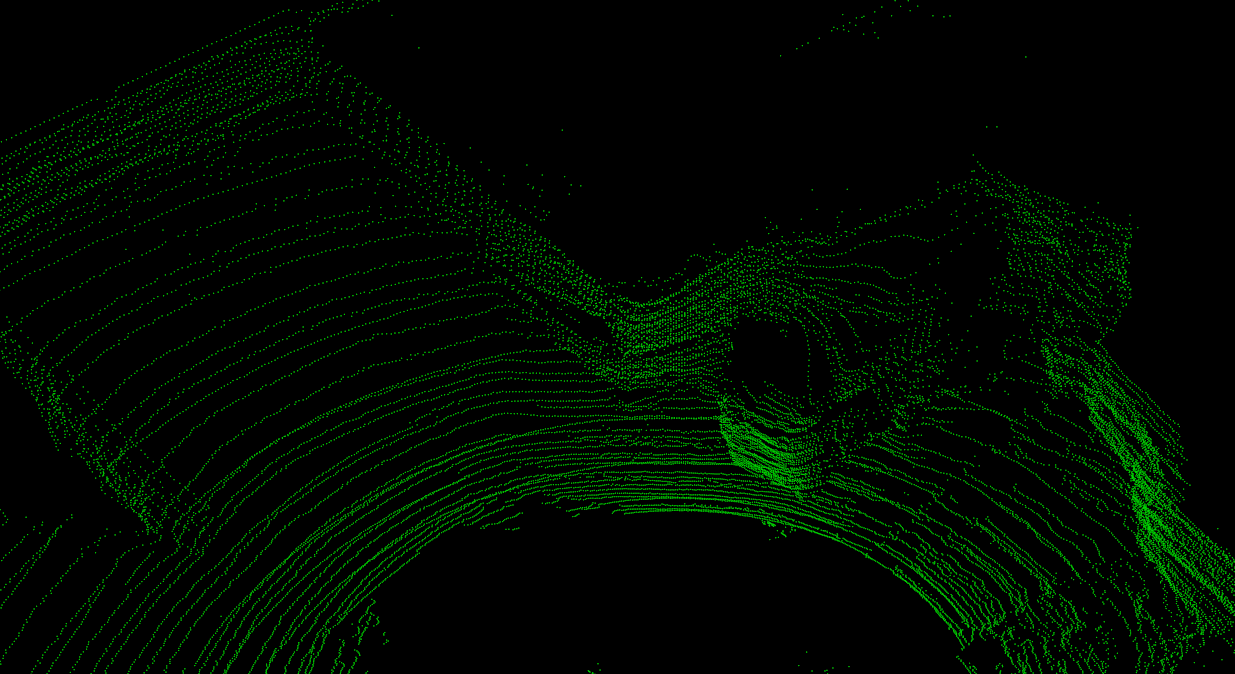}
        
    }

    \subfloat[][]
    {
        \includegraphics[width=0.45\textwidth]{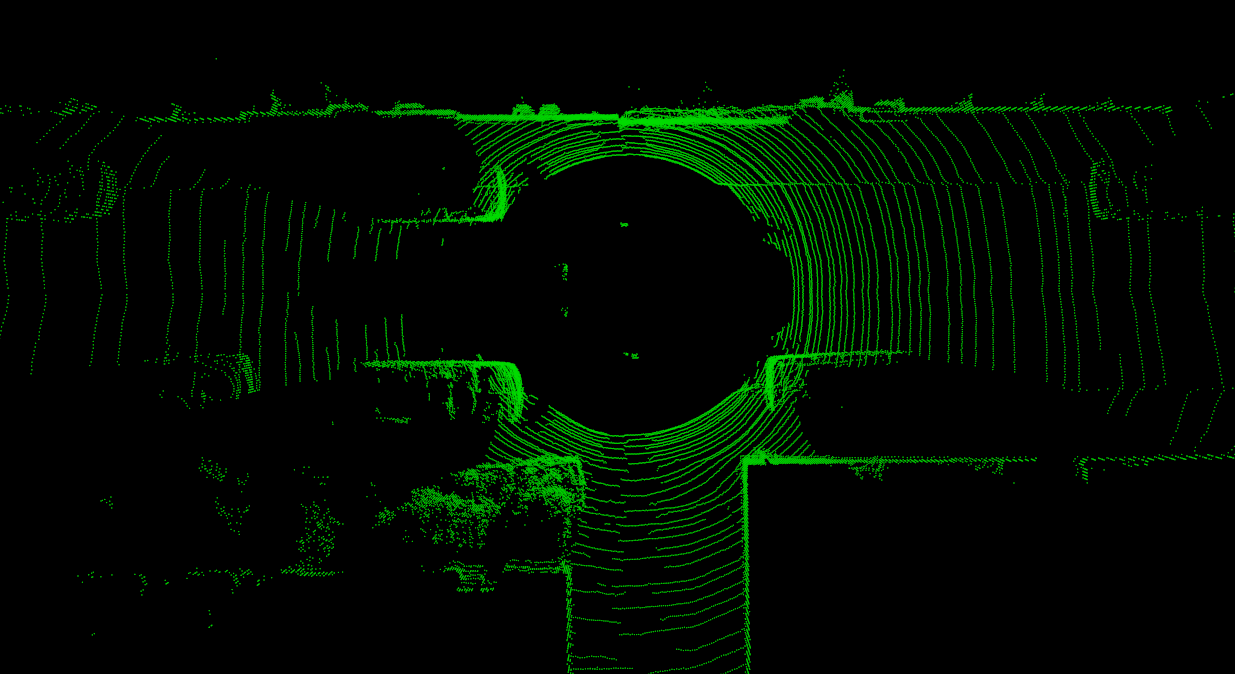}
        
    }\subfloat[][]
    {
        \includegraphics[width=0.45\textwidth]{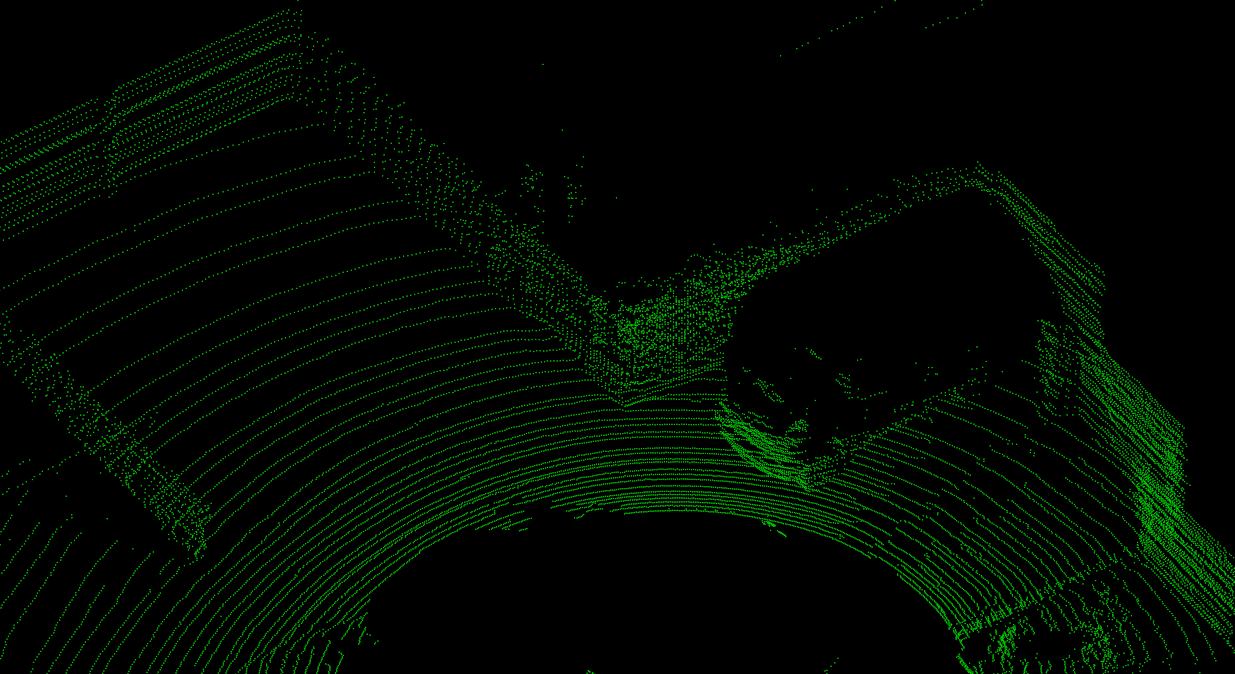}
        
    }
    \caption{Qualitative results for densification with and without simultaneous sampling (R2DM Base). Top to bottom: Input, Default Sampling, Simultaneous Sampling, Ground Truth. Our method has stronger geometric consistency, shown by cars and similar objects not merging with walls, and having less blurred sides.}
    \label{fig:DensificationImages}    
\end{figure}
\begin{figure}
    \centering
    \subfloat[][]
    {
        \includegraphics[width=0.45\textwidth]{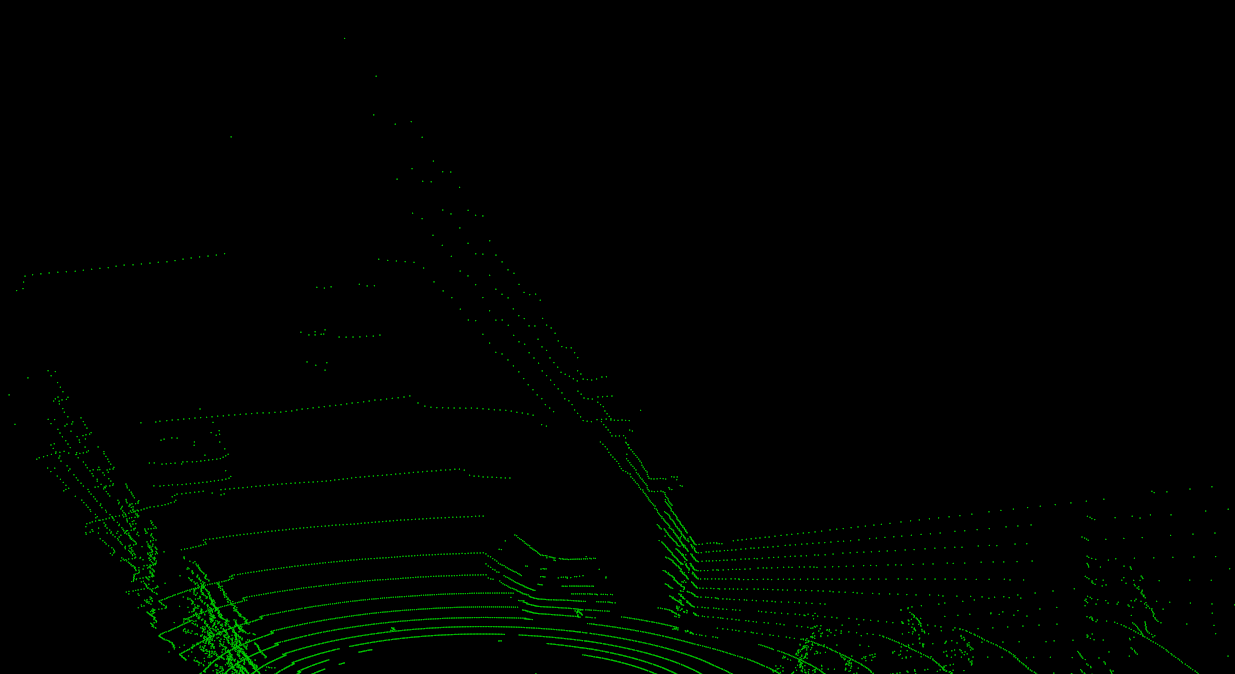}
        
    }
    \subfloat[][]
    {
        \includegraphics[width=0.45\textwidth]{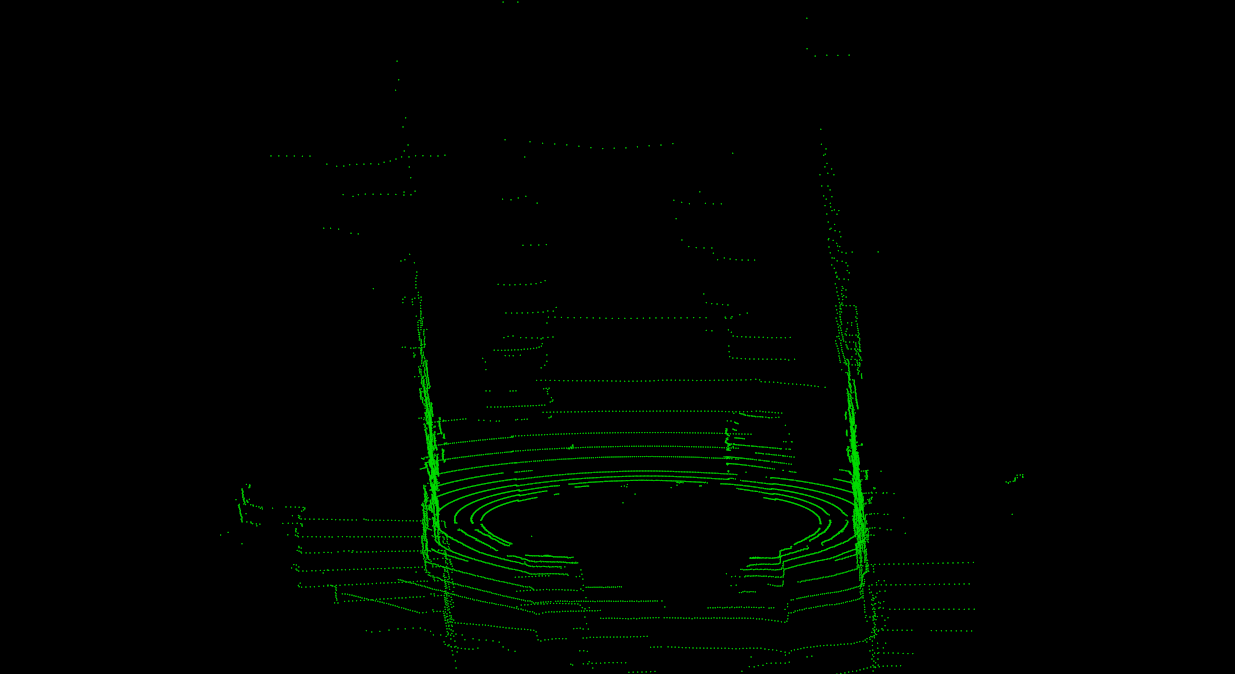}
        
    }\\
    \subfloat[][]
    {
        \includegraphics[width=0.45\textwidth]{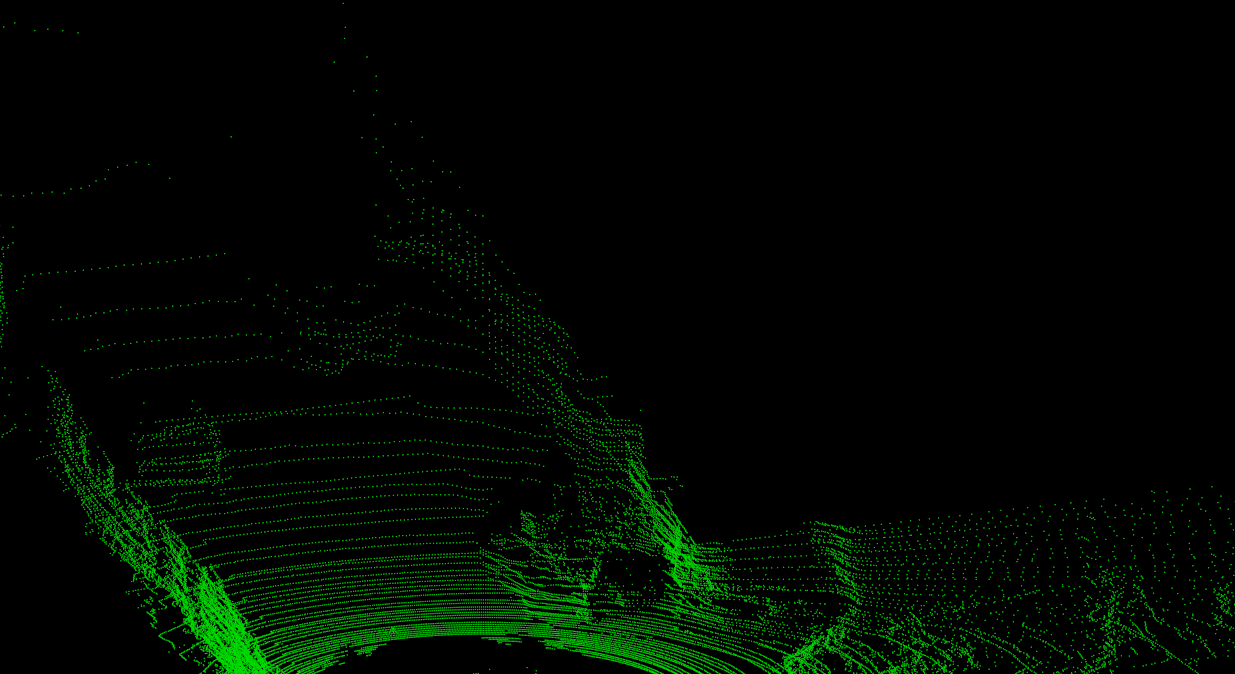}
        
    }
    \subfloat[][]
    {
        \includegraphics[width=0.45\textwidth]{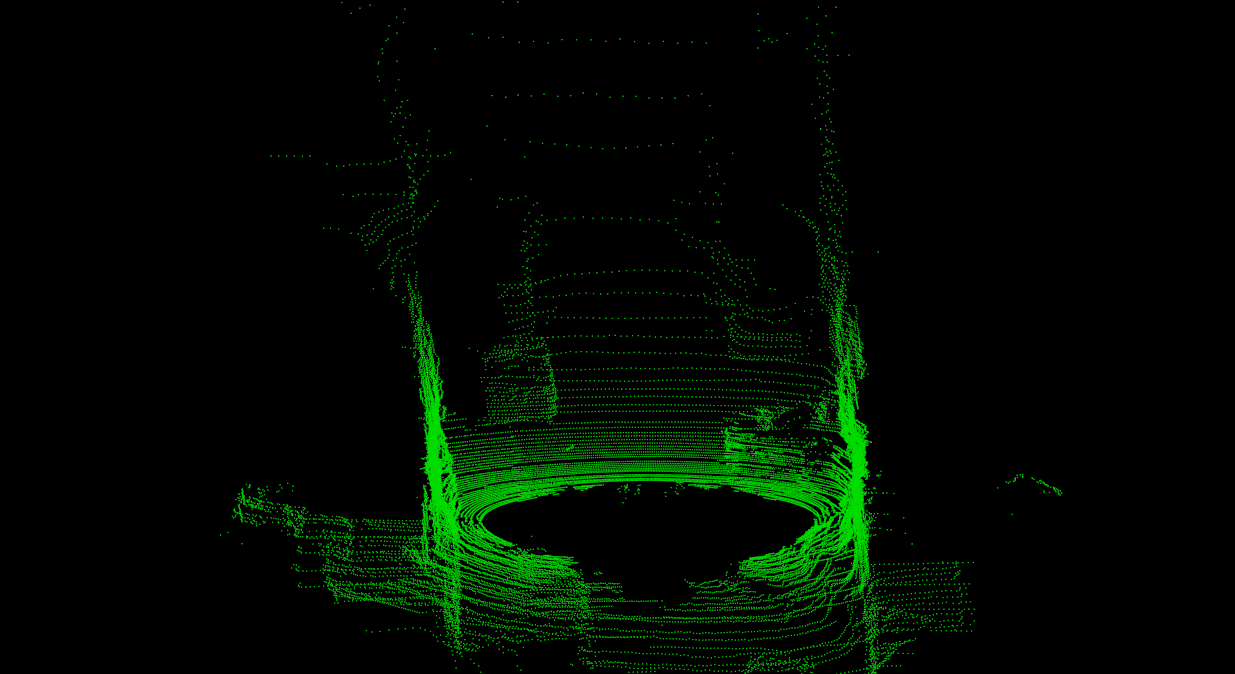}
        
    }\\
     \subfloat[][]
    {
        \includegraphics[width=0.45\textwidth]{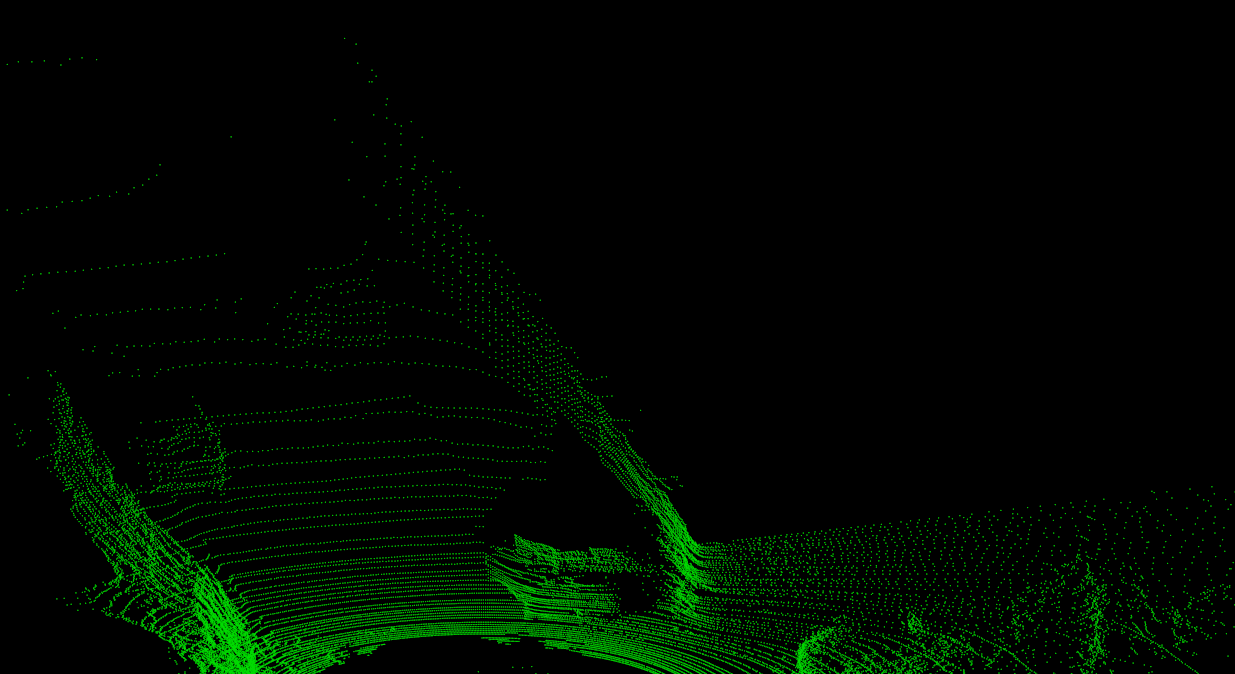}
        
    }
    \subfloat[][]
    {
        \includegraphics[width=0.45\textwidth]{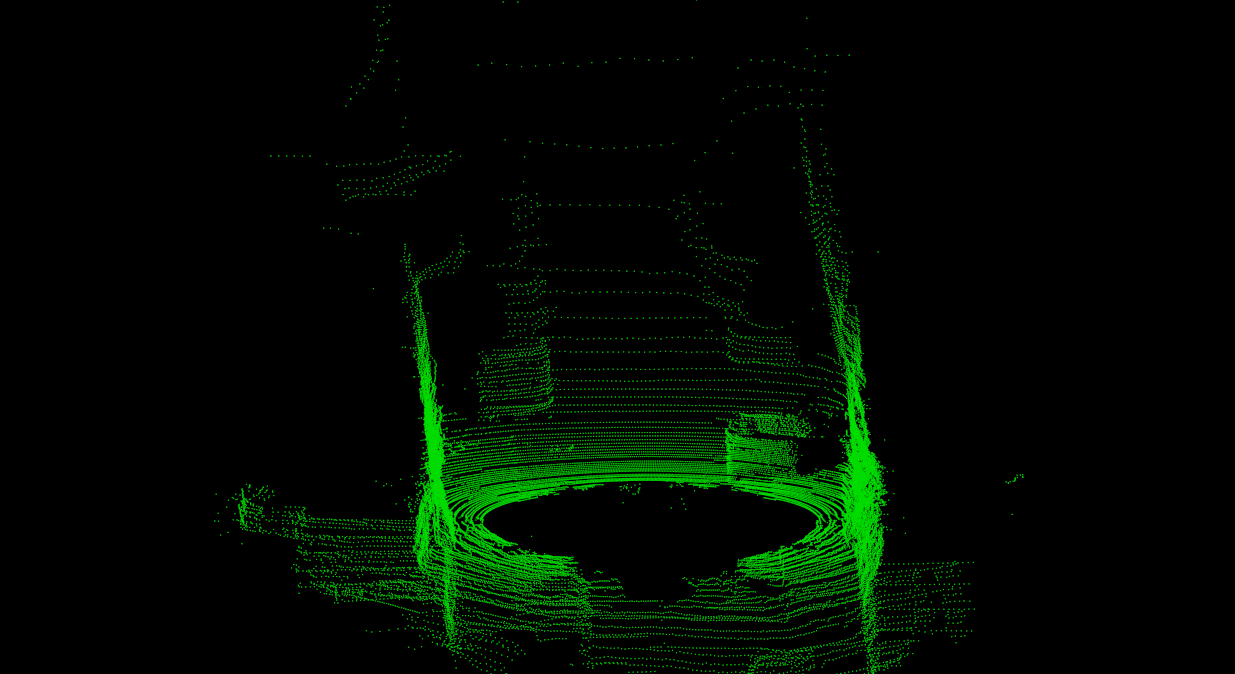}
        
    }

    \subfloat[][]
    {
        \includegraphics[width=0.45\textwidth]{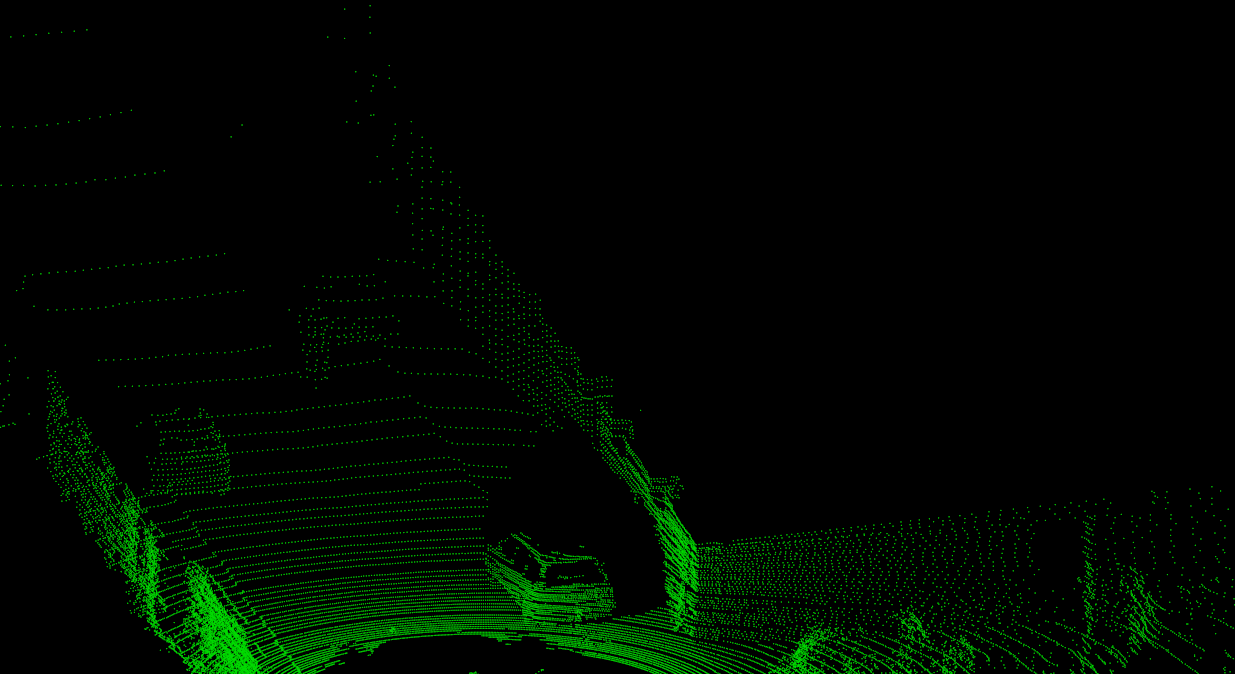}
        
    }\subfloat[][]
    {
        \includegraphics[width=0.45\textwidth]{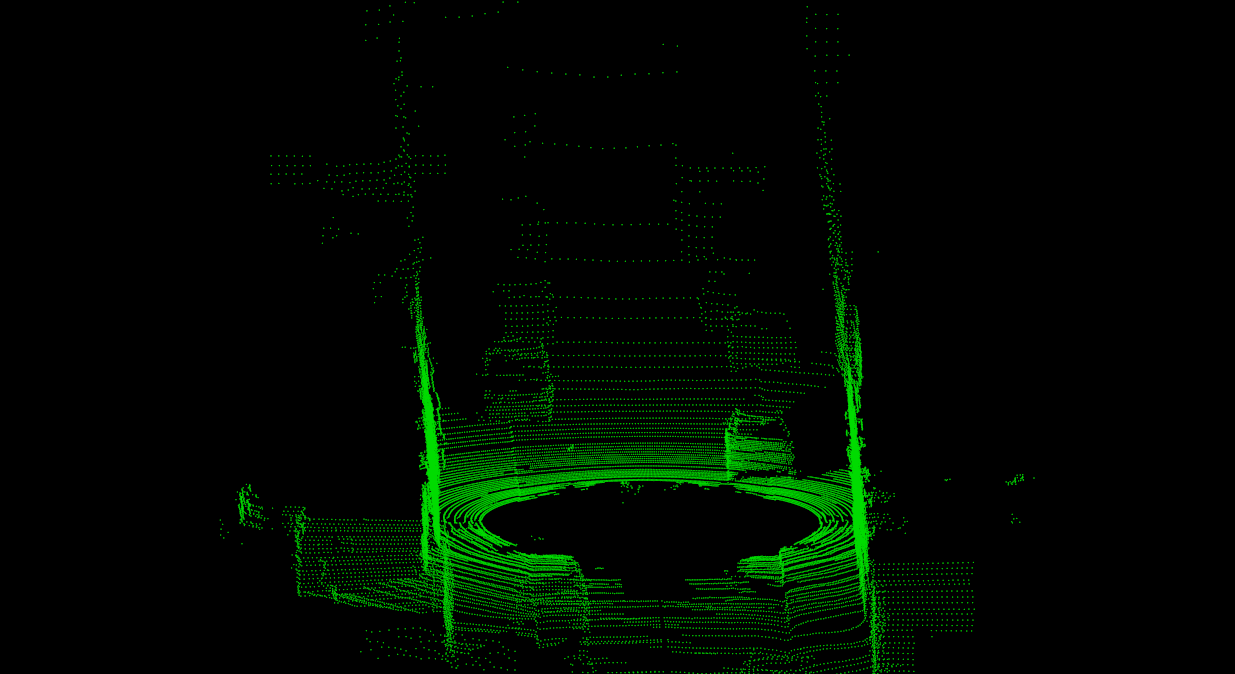}
        
    }
    \caption{More Qualitative results for densification with and without simultaneous sampling (R2DM Base). Top to bottom: Input, Default Sampling, Simultaneous Sampling, Ground Truth. Our simultaneous sampling shows clearer objects in the scene, with less noise and clearer edges between neighbouring objects}
    \label{fig:DensificationImagesTwo}    
\end{figure}

\begin{figure}
    \centering
    \subfloat[][]
    {
        \includegraphics[width=0.45\textwidth]{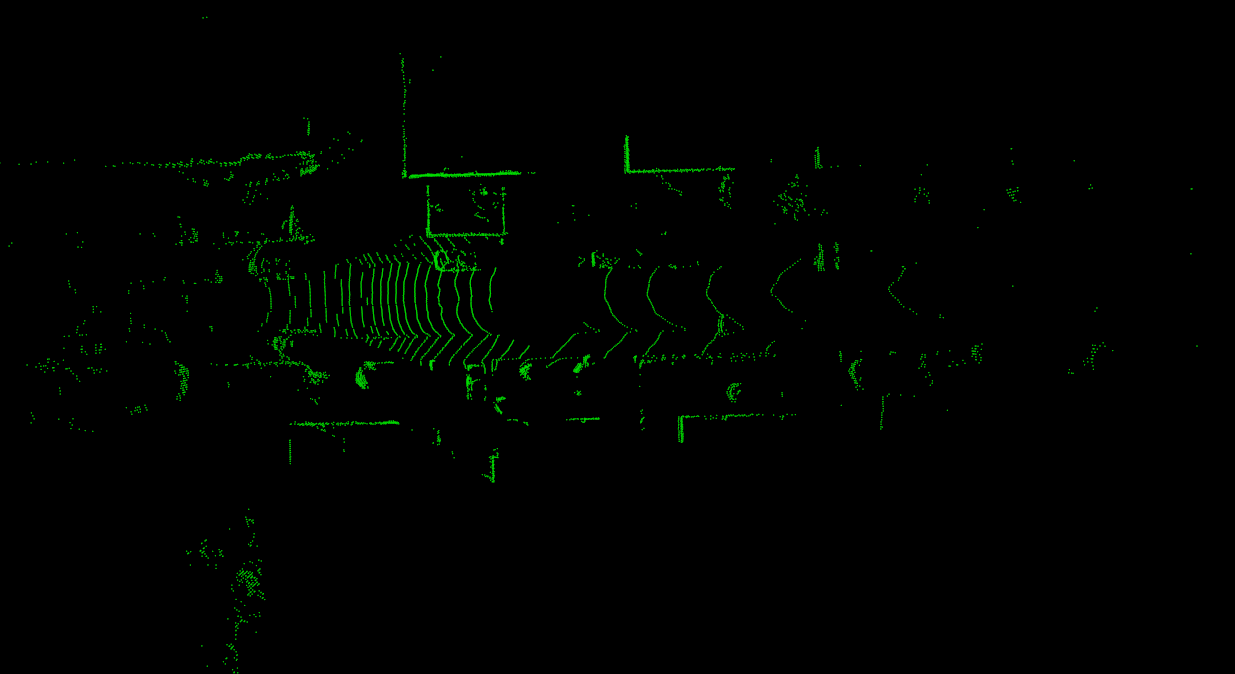}
        
    }
    \subfloat[][]
    {
        \includegraphics[width=0.45\textwidth]{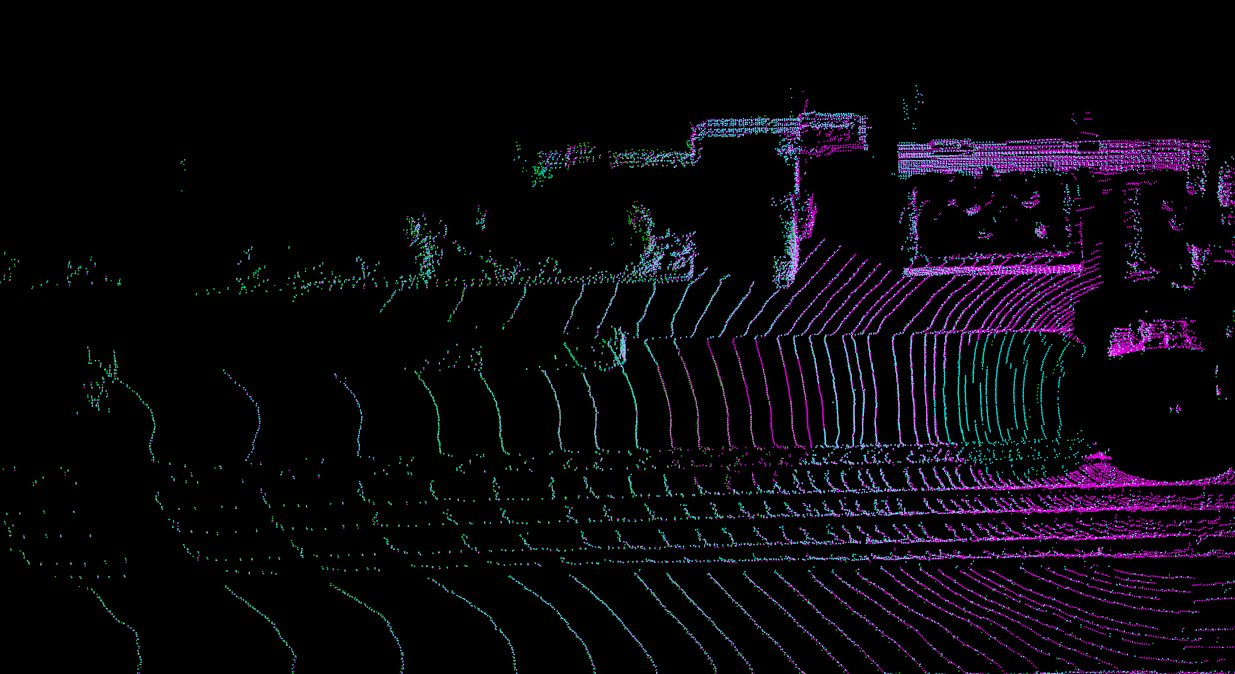}
        
    }\\
    \subfloat[][]
    {
        \includegraphics[width=0.45\textwidth]{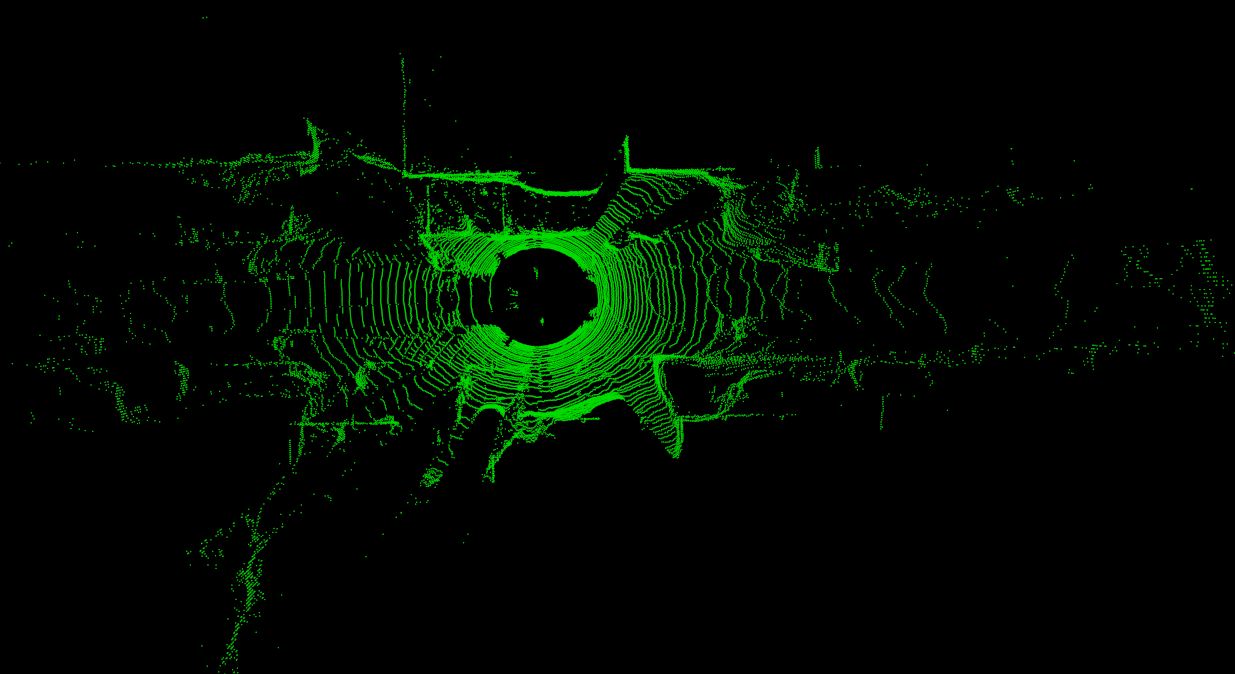}
        
    }
    \subfloat[][]
    {
        \includegraphics[width=0.45\textwidth]{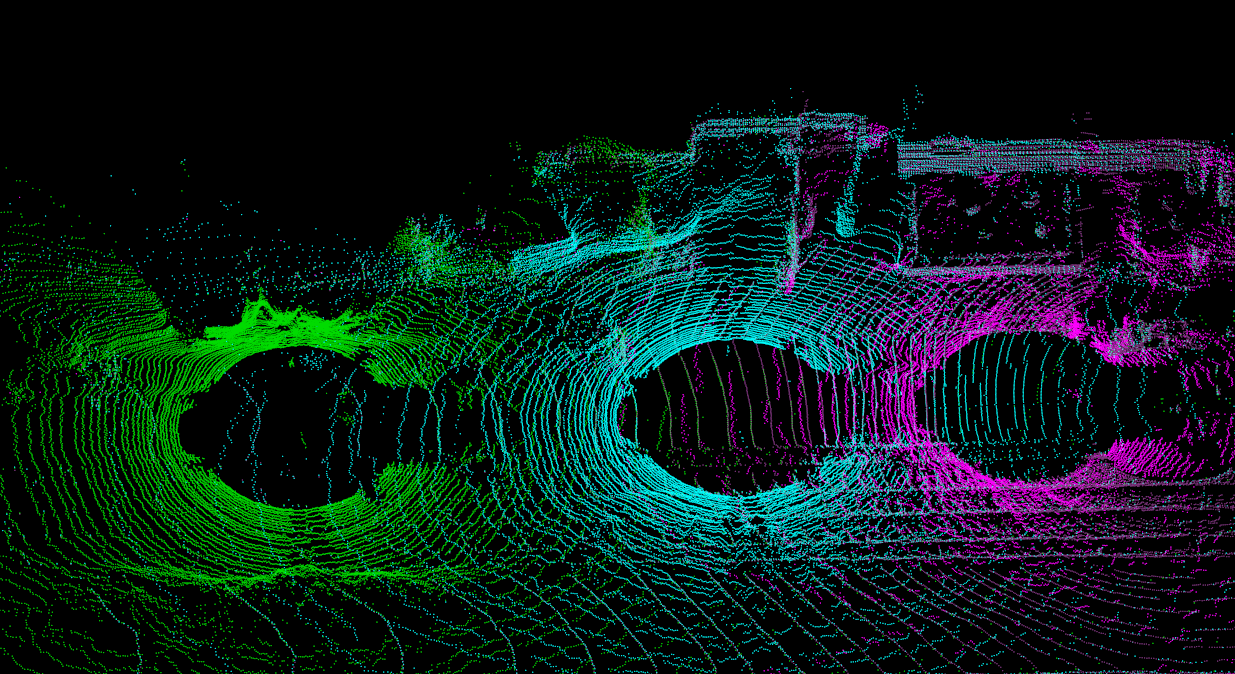}
        
    }\\
     \subfloat[][]
    {
        \includegraphics[width=0.45\textwidth]{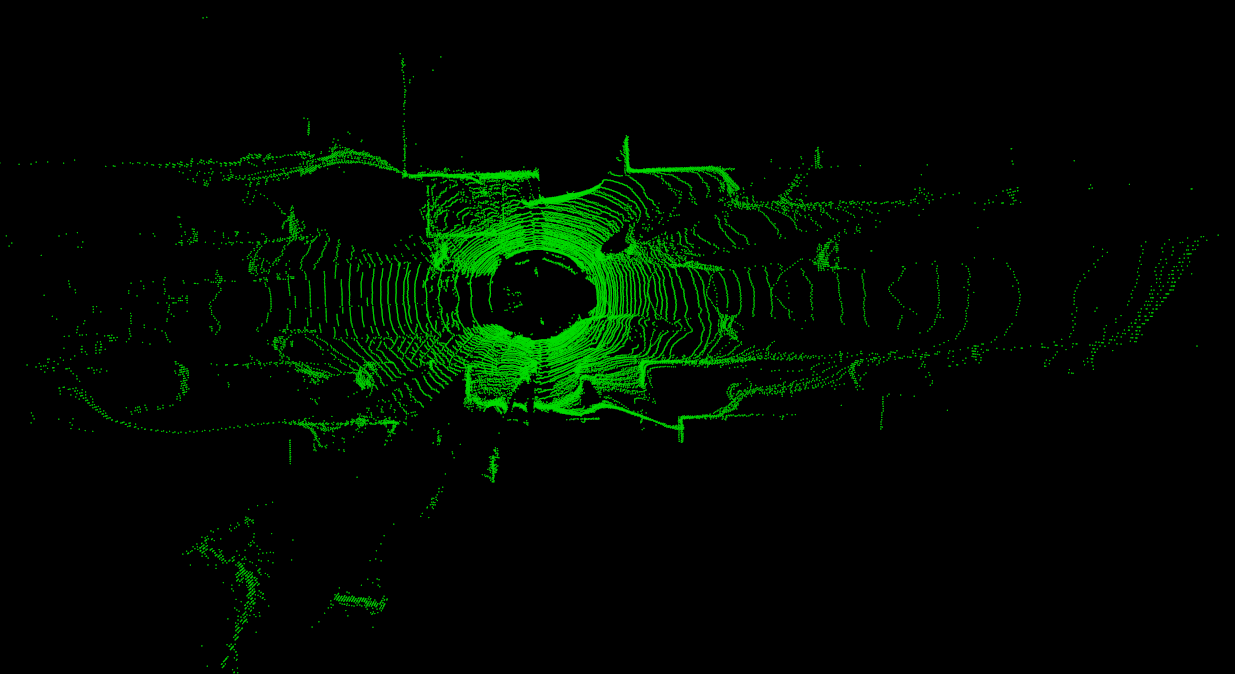}
        
    }
    \subfloat[][]
    {
        \includegraphics[width=0.45\textwidth]{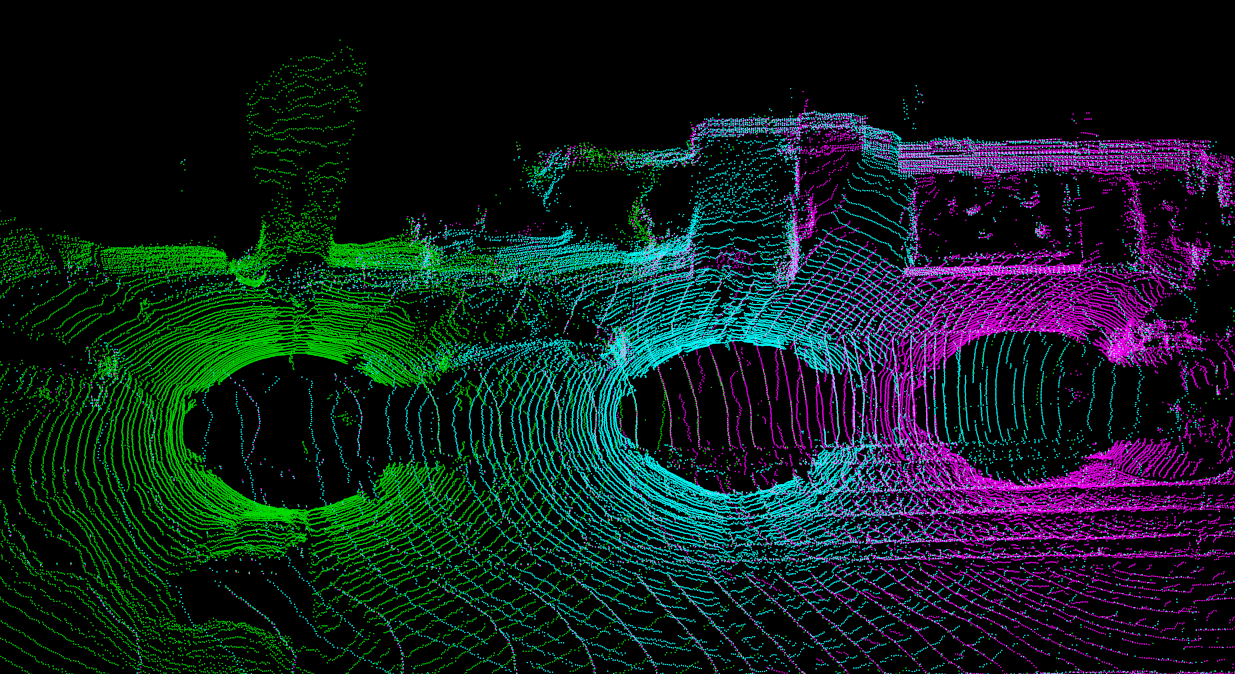}
        
    }

    \subfloat[][]
    {
        \includegraphics[width=0.45\textwidth]{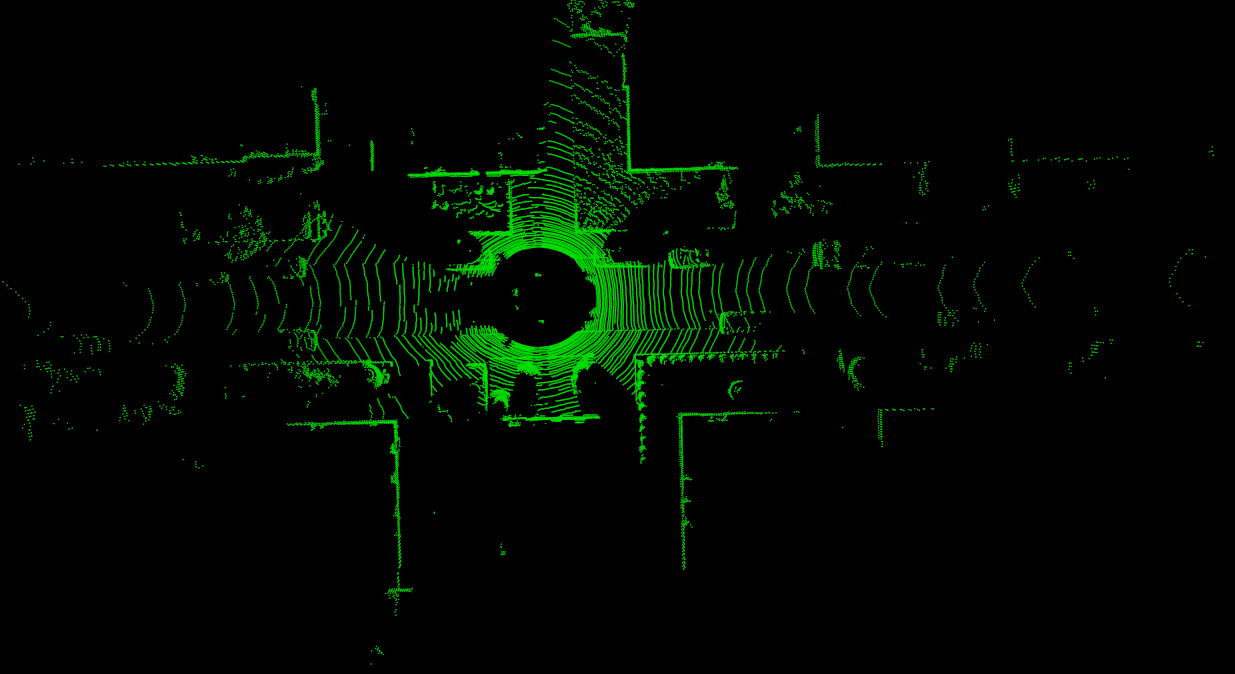}
        
    }\subfloat[][]
    {
        \includegraphics[width=0.45\textwidth]{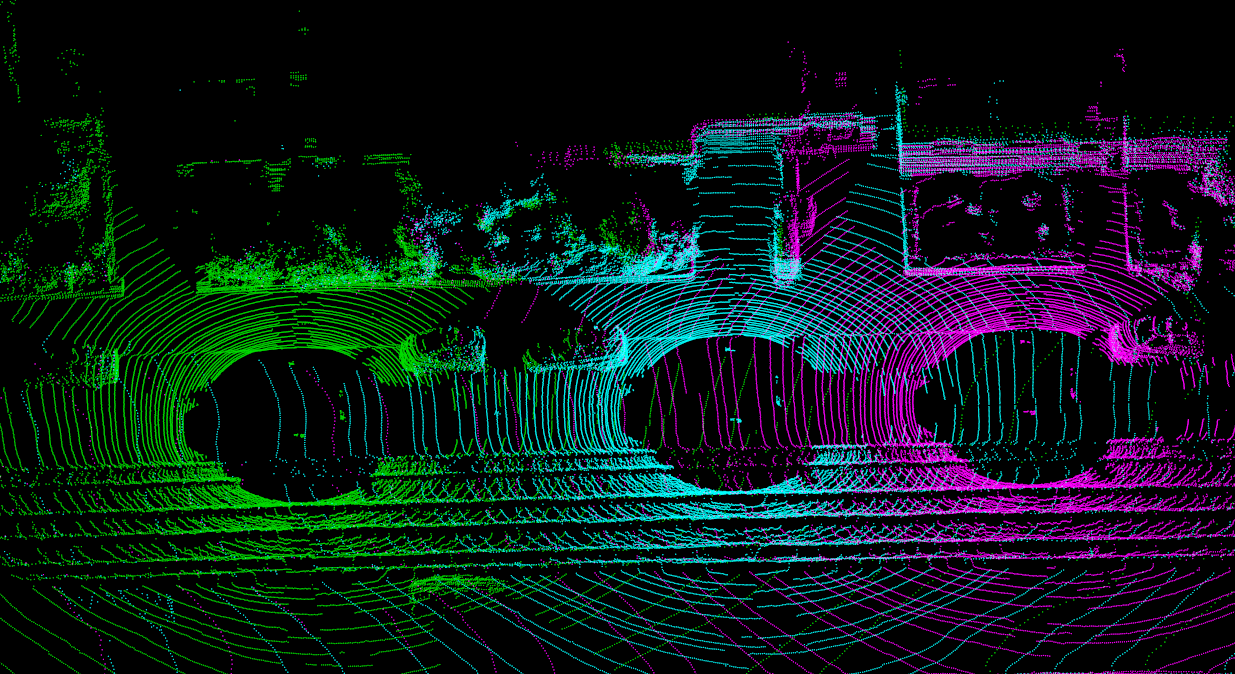}
        
    }
    \caption{Qualitative results for novel view generation with and without simultaneous sampling (R2DM Base). Top to bottom: Input, Default Sampling, Simultaneous Sampling, Ground Truth. Right column shows three novel views generated from the same input, coloured separately to aid visualisation. \newline Unlike densification, novel view generation requires generating novel geometry. With our simultaneous sampling, generated walls and objects are more consistent with both the input and the other synthetic scans generated from it (\eg simultaneous sampling's green scan has a wall which aligns with the blue scan, while in single view it's wall is too close to the green scan's origin, inconsistent with the scene.)}
    \label{fig:DensificationImagesTwo}    
\end{figure}
\end{document}